\documentclass[journal]{IEEEtran}

\ifCLASSINFOpdf
\else
\fi

\usepackage{times}
\usepackage{graphicx}
\usepackage{amsmath}
\usepackage{amssymb}
\usepackage{algorithm}
\usepackage{algorithmic}
\usepackage{multirow}
\usepackage{booktabs}
\usepackage{adjustbox}
\usepackage{subfigure}

\begin{document}

\title{Similarity Guided Deep Face Image Retrieval}

\author{Young~Kyun~Jang, and 
        Nam~IK~Cho,~\IEEEmembership{Senior Member, IEEE}
\thanks{The authors are with the Department of Electrical and Computer Engineering, Seoul National University, Seoul 151-742, South Korea, and also with INMC, Seoul 151-742, South Korea (e-mail: kyun0914@ispl.snu.ac.kr; nicho@snu.ac.kr)}}

\markboth{Journal of \LaTeX\ Class Files,~Vol.~14, No.~8, August~2015}%
{Shell \MakeLowercase{\textit{et al.}}: Bare Demo of IEEEtran.cls for IEEE Journals}

\maketitle

\begin{abstract}

Face image retrieval, which searches for images of the same identity from the query input face image, is drawing more attention as the size of the image database increases rapidly. In order to conduct fast and accurate retrieval, a compact hash code-based methods have been proposed, and recently, deep face image hashing methods with supervised classification training have shown outstanding performance. However, classification-based scheme has a disadvantage in that it cannot reveal complex similarities between face images into the hash code learning. In this paper, we attempt to improve the face image retrieval quality by proposing a Similarity Guided Hashing (SGH) method, which gently considers self and pairwise-similarity simultaneously. SGH employs various data augmentations designed to explore elaborate similarities between face images, solving both intra and inter identity-wise difficulties. Extensive experimental results on the protocols with existing benchmarks and an additionally proposed large scale higher resolution face image dataset demonstrate that our SGH delivers state-of-the-art retrieval performance.
\end{abstract}


\IEEEpeerreviewmaketitle

\section{Introduction}
\label{sec:1}
\IEEEPARstart{L}{earning} to hash for image retrieval has made significant progress since the introduction of deep learning. In particular, many researchers attempted to produce compact binary hash codes through supervised learning with image class labels \cite{DSH, DHN, DSDH, HashNet, CBH}, which are shown to provide superior performance to existing methods. Face image retrieval is also being advanced using deep learning \cite{DHCQ, DDH, DDQH, DCBH, DAGH} to learn facial representations of each identity.

Face image retrieval can be considered a sub-problem of general image retrieval, finding images of the same identity (class label) to the query. Still, face image retrieval has subtle differences from the general task due to some particular properties of facial image datasets. To be specific, 1) the variance of the data distribution within a class is often high due to makeup, facial expression, glasses, view-points, etc. (intra-identity difficulties), 2) the similarity between the two different classes is comparatively high because there are many similar faces (inter-identity resemblance), and 3) there are relatively many classes (identities) to distinguish. Therefore, retrieval systems for face image hashing attempt to improve performance by increasing discriminativity among binary hash codes assigned to each identity.

For this purpose, existing deep face image retrieval methods usually employ common classification losses such as softmax cross-entropy, and utilize the intermediate feature vector as a hash code. Although this approach provides moderate retrieval performance, there exists a limitation in representing the semantic similarity between images because the hash codes are learned only with discrete target labels (0 and 1 if one-hot encoded). Therefore, the retrieval performance can be improved if the similarities between face images are elaborately considered in hash function learning.

Inspired from the works of the self-supervised metric learning \cite{self_survey, S4l, Simclr, Sup_Cont, Curl}, which has been widely explored to understand diverse image content representation, we aim to find similarity between images by contrasting them. First, a series of data augmentation techniques is employed to obtain the transformed image. Then, this image is utilized to define the semantic similarities among images. To be specific, transformed images originating from the same image are considered similar (self-similarity), and the other transformed images from different images are considered dissimilar (pairwise-similarity). In our case, we exploit the allocated label information when determining the pairwise-similarity, and the transformed images that originate from the different images of the same identity are regarded as positive pairs.

\begin{figure}[!t]
\centering
\includegraphics[width=0.99\linewidth]{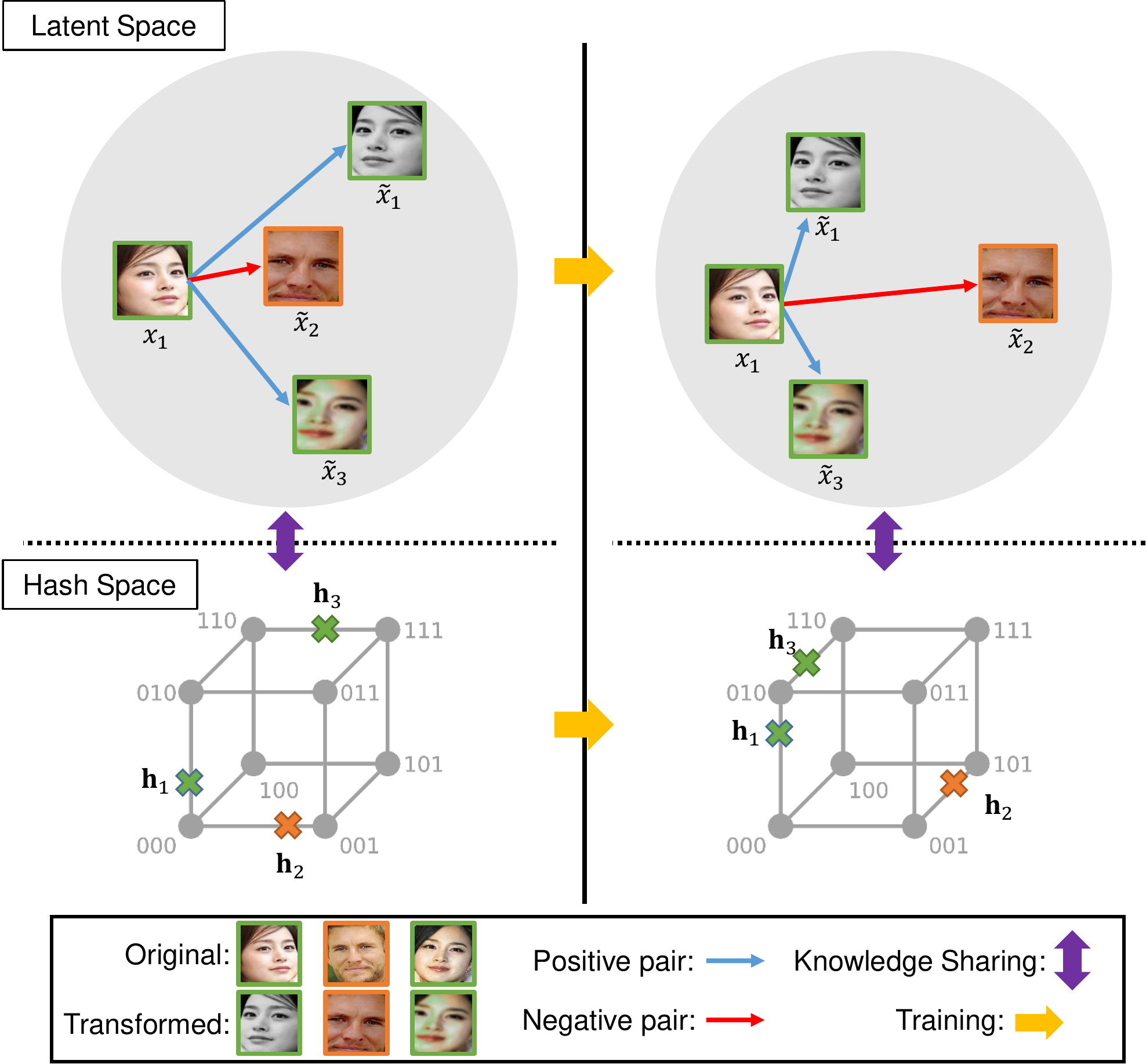}
\caption{A simple visualization of our proposed method. $\tilde{x}_{1},\tilde{x}_{2},\tilde{x}_{3}$ are the transformed output of $x_{1},x_{2},x_{3}$, respectively, and we only illustrate the case of $x_{1}$ for simplicity. $x_{1}$ forms a positive pair with $\tilde{x}_1$, and $\tilde{x}_{3}$ derived from the same identity, and forms a negative pair with $\tilde{x}_2$ derived from the different identity. With this similarity information learned in the latent space, the final output hash codes $\mathbf{h}_{1},\mathbf{h}_{2},\mathbf{h}_{3}$ which are originated from $x_{1},x_{2},x_{3}$, respectively, can preserve self and pairwise-similarity near the Hamming space (cubic), designed to accomplish high retrieval scores.}

\label{fig:Figure1}
\end{figure}

For better understanding, we illustrate our conceptual scheme in Figure \ref{fig:Figure1}. The combination of random data augmentations such as crop and resize, flip, color distortion, and Gaussian blurring is applied to face images with the consideration of facial properties. The self-similarity and the pairwise-similarities are considered simultaneously in the latent space during training. At last, the learned similarity knowledge is shared between the latent space and the hash space to obtain the discriminative binary hash codes.

In this paper, we propose the first self and pairwise-similarity assisted convolutional neural network-based hashing framework; named $\textit{Similarity Guided Hashing}$ (SGH), which generates binary-like hash codes for fast and accurate retrieval. Specifically, we introduce two novel training objectives. First, a novel Similarity Pairing loss is introduced on transformed face images with human-supervised information (identity label annotations). This approach allows a more sophisticated understanding of the correlations between images. Second, to alleviate the gap between discrete binary code and the hash code, we suggest a squared Quantization loss to minimize it fast and moderately. Additionally, a $l2$-regularization is utilized to avoid severe deviations between the original and the transformed image feature representations, and a standard Cross-Entropy function for classification is employed to enhance the general distinctiveness between hash codes. The entire loss functions are applied to SGH in an end-to-end manner, without any additional complex training batch configuration strategy.

To demonstrate that our method excels previous methods in various conditions, we construct a new experimental protocol using a face detection algorithm (DSFD) \cite{DSFD} on VGGFace2 \cite{VGGFace2} test set, a large scale image dataset for face recognition. The dataset we configure holds three times higher resolution images than the existing protocols, thus containing richer facial feature representations. Furthermore, we conduct identity-disjoint experiments proposed in \cite{DAGH} with the subset of VGGFace2 training set. The results show that our SGH encodes face images into the Hamming space properly, despite the absence of the identity information for the network training.

We can summarize our contributions as follows:

\begin{itemize}
\item To the best of our knowledge, this is the first work that employs elaborate similarity learning in the latent space to guide the deep hashing-based image retrieval system training.

\item The discriminative binary-like hash codes are directly obtained to conduct fast and accurate face image retrieval in an end-to-end framework, where the learned self and pairwise-similarity knowledge is well presented.

\item Extensive experimental results on both low and high-resolution face image datasets demonstrate that our SGH yields the state-of-the-art performance for the benchmark protocols.
\end{itemize}

\section{Related Works}
\label{sec:2}

This section presents a brief introduction to the deep hashing methods for general images and face images. Refer to the survey \cite{Survey_Hash} to see the early works in non-deep learning binary hashing (ITQ \cite{ITQ}, SH \cite{SH}, KSH \cite{KSH}, SDH \cite{SDH}) which can be utilize as hashing system for face image retrieval.

\subsection{General image hashing methods}
Convolutional Neural Network (CNN)-based hashing approaches with fully supervised learning ~\cite{CNNH, DSH, CSQ} are leading the mainstream with promising outcomes. For example, CNN Hashing (CNNH)~\cite{CNNH} utilizes a CNN to generate compact hash codes by training the network with the given pairwise label information. Deep Supervised Hashing (DSH) ~\cite{DSH} learns hash codes by approximating discrete values with relaxation and training them with the supervised signals. Most recently, the method that exploits the global similarity metric \cite{CSQ} shows outstanding performance in the supervised scheme. Meanwhile, for hashing with few labels or without labels, deep semi-supervised learning \cite{GPQ} and deep unsupervised learning \cite{TBH} also attracted attention with high scores. 

\subsection{Face image hashing methods}
In terms of face images, there have been several CNN-based hashing approaches \cite{DHCQ,DDH,DDQH,DCBH,DAGH} that take into account face characteristics. In \cite{DHCQ}, they proposed an end-to-end framework that simultaneously learns face features and the binary hash codes by minimizing classification and quantization loss at once, named Deep Hashing based on Classification and Quantization errors (DHCQ). Discriminative Deep Hashing (DDH) method \cite{DDH} integrated divide-and-encode modules to reduce the redundancies among hash codes and the network parameters to improve DHCQ. Discriminative Deep Quantization Hashing (DDQH) \cite{DDQH} upgraded DDH by expanding the number of channels in layers, replacing the divide-and-encode module with a fully-connected layer, and applied a batch normalization quantization module. Discrete Attention Guided Hashing (DAGH) \cite{DAGH} employed discrete identity loss and a multi-attention cascade network architecture to capture face features.

Our SGH is the first work to split the latent space and the hash space, and introduce a self and pairwise-similarity learning. By assisting deep hashing model training with complicated similarity knowledge, SGH is able to surpass existing algorithms with state-of-the-art experimental outcomes. Furthermore, especially to precisely assess the advantage of our proposal in face image retrieval protocols, we configure a new dataset with larger images than the previous ones, and SGH also shows the best results.

\section{Methodology}
\label{sec:3}

\begin{figure*}[!t]
\centering
\includegraphics[width=0.99\linewidth]{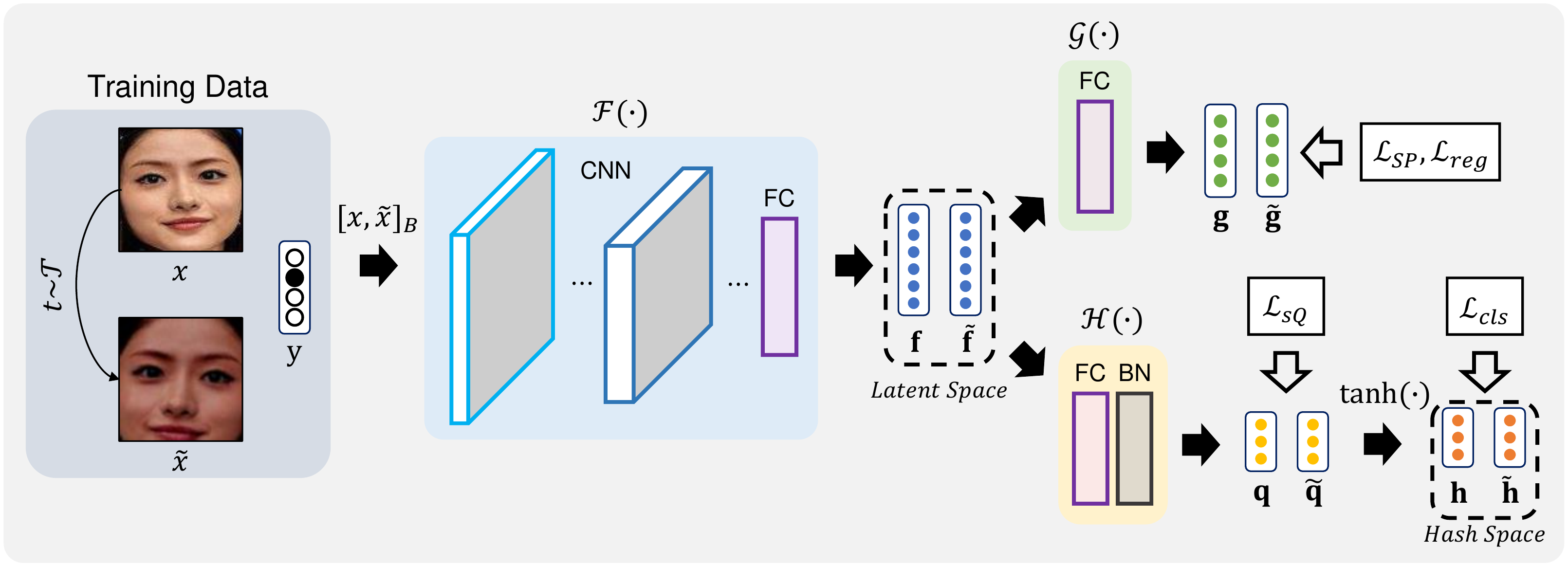}
\caption{The overall training process of SGH. Training data $x$ and its corresponding one-hot encoded label $\mathrm{y}$ is prepared with $\tilde{x}$, which is the output of the combination of several data augmentation techniques. The concatenation operation $[\cdot]_B$ along the batch dimension is applied to $x$ and $\tilde{x}$ before entering the framework. There are three trainable components: feature extractor $\mathcal{F}(\cdot)$, projection head $\mathcal{G}(\cdot)$, and hashing head $\mathcal{H}(\cdot)$. The entire framework is trained end-to-end with four loss functions; 1) Similarity Pairing loss $\mathcal{L}_{SP}$ which simultaneously learns self-similarity and pairwise-similarity to understand diverse image representations, 2) $l2$-regularization $\mathcal{L}_{reg}$ to diminish deviations between the original and the transformed representations, 3) squared Quantization loss $\mathcal{L}_{sQ}$ to reduce the gap between the discrete binary code and the hash code, and 4) classification loss $\mathcal{L}_{cls}$ to increase the general discriminability between hash codes. The learned similarity knowledge is shared in the latent space of $\mathbf{f}$ and $\mathbf{\tilde{f}}$. In the retrieval phase, $\mathcal{F}(\cdot)$ and $\mathcal{H}(\cdot)$ are used to generate hash code $\mathrm{h}$, and the output binary code $\mathbf{b}$ is obtained with the sign function.}

\label{fig:Figure2}
\end{figure*}

The goal of deep hashing for face image retrieval is to map an input image $x$ to an identity-wise discriminative $K$-bits binary code $\mathbf{b} \in \{-1,1\}^K$. However, the network training process does not include binarization, since the sign function used for binary encoding is non-differentiable. Instead, our deep model is trained in the real-valued space with the self-similarity and pairwise-similarity to learn continuous facial representations. At the same time, we minimize the quantization error that occurs during the binary conversion to reduce the gap between discrete binary code and real-valued hash code. Also, the knowledge learned in the latent space is shifted into the hash space while minimizing the classification error of hash codes to increase retrieval accuracy.

\begin{table}[!ht]
\centering
\caption{Detailed network architecture of SGH. The output size is described according to the input image resolution.}
\begin{adjustbox}{width=0.49\textwidth}
\begin{tabular}{cccc}
\toprule 
\multirow{2}{*}{Name} & Output Size & Output Size & \multirow{2}{*}{Layer}\\
 & $(Small)$ & $(Large)$ & \\
\midrule
conv 1 & $32\times32\times64$& $96\times96\times64$ & $3\times3$, stride 1\\
\midrule
layer 1 &  $32\times32\times64$& $96\times96\times64$ & \Bigg[ \begin{tabular}{@{}c@{}}$3\times3, 64$ \\ $3\times3, 64$ \end{tabular} \Bigg] $\times 2$\\
\midrule
layer 2 & $16\times16\times128$& $48\times48\times128$ & \Bigg[ \begin{tabular}{@{}c@{}}$3\times3, 128$ \\ $3\times3, 128$ \end{tabular} \Bigg] $\times 2$\\
\midrule
layer 3 & $8\times8\times256$& $24\times24\times256$ & \Bigg[ \begin{tabular}{@{}c@{}}$3\times3, 256$ \\ $3\times3, 256$ \end{tabular} \Bigg] $\times 2$\\
\midrule
layer 4 & $4\times4\times512$& $12\times12\times512$ & \Bigg[ \begin{tabular}{@{}c@{}}$3\times3, 512$ \\ $3\times3, 512$ \end{tabular} \Bigg] $\times 2$\\
\midrule
pool & $1\times1\times512$ & $1\times1\times512$& average pool\\
\midrule
$\text{FC}_{latent}$ & $512$ & $512$ & $512\times512$ fully connect.\\
\midrule
$\text{FC}_{proj}$ & $128$ & $128$ & $512\times128$ fully connect.\\
\midrule
$\text{FC}_{hash}$ & \# Bits & \# Bits & $512\times$ \# Bits fully connect.\\
\midrule
$\text{FC}_{cls}$ & \# IDs & \# IDs & \# Bits $\times$ \# IDs fully connect.\\
\bottomrule
\end{tabular}
\end{adjustbox}
\label{table:Table1}
\end{table}

\subsection{Framework}

As illustrated in Figure \ref{fig:Figure2}, our Similarity Guided Hashing (SGH) consists of three components with trainable parameters. From a face image input x, feature extractor $\mathcal{F}(\cdot)$ outputs the face feature vector $\mathbf{f}$. Any common CNN architecture such as AlexNet \cite{AlexNet}, VGG-F \cite{CNN-F} or ResNet \cite{ResNet} can be employed as $\mathcal{F}(\cdot)$. We select ResNet18 as our baseline with a simple modification in the number of convolutional filter channels. In order to obtain the hash code of the desired length and conduct latent space training, additional FC layers are utilized. The detailed network architecture is listed in the Table \ref{table:Table1}.

Following the observations in \cite{Simclr, Sup_Cont}, we introduce an additional projection head $\mathcal{G}(\cdot)$ to learn the self and pairwise-similarity with the dimension-reduced vector $\mathbf{g}$, rather than directly utilizing $\mathbf{f}$ for training. Note that, we only apply a single fully-connected layer (FC) without two or more FC layers with non-linearity operations. This has the advantage of being able to transfer the self and pairwise-similarity knowledge learned from $\mathbf{g}$ to $\mathbf{f}$ without redundancy.

Hashing head $\mathcal{H}(\cdot)$ is similar to the BNQ module proposed in \cite{DDQH}. The FC layer embeds the latent space into the hash space of the desired bit length, and the batch normalization layer \cite{BatchNorm} (BN) speeds up the convergence and improves the stability of learning. The output of the hashing head, $\mathbf{q}$, is mapped to $\mathbf{h}$ consisting of real values between -1 and 1 through the $\mathrm{tanh}(\cdot)$ function. When we conduct retrieval, the input image x is converted to the hash code $\mathbf{h}$ by passing through the SGH framework, and finally becomes a discrete binary code with the sign function.

\subsection{Training}

\begin{figure}[!t]
\centering
\includegraphics[width=0.99\linewidth]{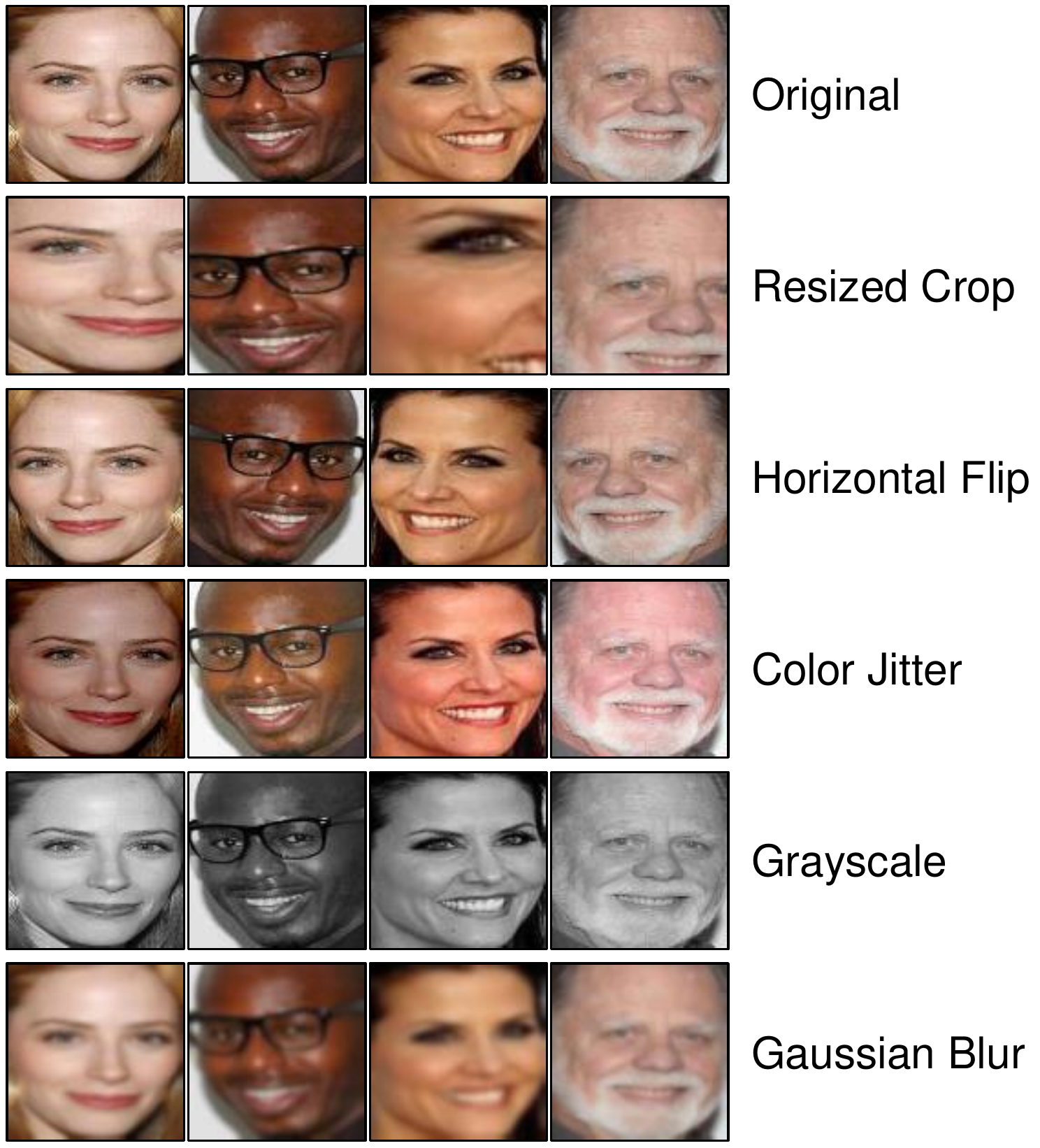}
\caption{Examples of face images in VGGFace2 dataset with various augmentation techniques applied.}

\label{fig:Figure3}
\end{figure}

In terms of data preparation, we follow the works presented in \cite{Simclr, Sup_Cont}, and apply various data augmentation techniques on the face images before training. As illustrated in Figure \ref{fig:Figure3}, we choose five methods to transform images and configure the data augmentation family $\mathcal{T}$. Most of the hyperparameters for augmentation are directly taken from \cite{Simclr} except for color jitter strength as 1/5, to take into account the skin color characteristics between different races. As listed in Figure \ref{fig:Figure3}, all transformations are applied with random probabilities in a sequential manner, starting with the resized crop and ending with the Gaussian blur. The augmented outputs are exploited to investigate the various underlying similarities of both self and other image pairs.

Suppose we are given a dataset with $N$ training samples; $\mathcal{X}=\{(x_1, \mathbf{y}_1),...,(x_N, \mathbf{y}_N)\}$ where each image $x_i$ is assigned a one-hot encoded label $\mathbf{y}_i\in \{0,1\}^c$ of $c$ identities. Transformed images are obtained as $\tilde{x}_i=t_i(x_i)$ for all samples $\{\tilde{x}_i\}^{N}_{i=1}$, where $t_i$ is sampled from $t\sim\mathcal{T}$. We propose four training objectives that learn complex facial representations in a high-dimensional latent space and make the hash code discriminative near the Hamming space with regularization. To cope with difficult-to-distinguish features of the face images such as similar appearance among different people, and variations in one person (facial expression, makeup, pose, illumination), we introduce Similarity Pairing loss that trains $\mathcal{F}(\cdot)$ and $\mathcal{G}(\cdot)$ as:

\begin{align}
\mathcal{L}_{SP}(\mathbf{g},\mathbf{y})=\frac{1}{N_B}\sum_{i=1}^{N_B}\left(\mathcal{L}_{CE}(S_i, Y_i)\right)
\label{equation:eqn1}
\end{align}

\noindent where $N_B$ denotes the number of images in a training batch. Similar to the concept proposed in \cite{NPair}, to conduct metric learning, we employ a standard cross-entropy loss $\mathcal{L}_{CE}$, where $S_i$ denotes dot similarity between the $i$-th $\mathbf{g}$ and every $\mathbf{\tilde{g}}$ in a batch; $S_i = \left[\mathbf{g}_i^T\mathbf{\tilde{g}}_1,...,\mathbf{g}_i^T\mathbf{\tilde{g}}_{N_B}\right]$, and $Y_i$ denotes a similarity between the $i$-th label and every label in a batch; $Y_i = \left[\mathbf{y}_i^T\mathbf{y}_1,...,\mathbf{y}_i^T\mathbf{y}_{N_B}\right]$. Notably, in order to balance the contribution, we apply $l1$-normalization on the label as $Y_i=Y_i/\lVert Y_i\rVert_1$. During training with $\mathcal{L}_{SP}$, the self-similarity part $\mathbf{g}_i^T\mathbf{\tilde{g}}_i$ diminishes the intra-identity difficulties, and the remaining pairwise-similarity parts contribute to increase the inter-identity discriminability.

As a means of regularization, additional $l2$ constraint is imposed on the embedding vectors as:

\begin{align}
\mathcal{L}_{reg}(\mathbf{g})=\frac{1}{N_B\cdot D_\mathbf{g}}\sum_{i=1}^{N_B}\sum_{j=1}^{D_\mathbf{g}}\left(g_{ij}^2\right)
\label{equation:eqn2}
\end{align}

\noindent where $D_\mathbf{g}$ is the dimensionality of $\mathbf{g}$, and $g_{ij}$ are the $j$-th element of $\mathbf{g}_i$. This regularization term forces the norm of $\mathbf{g}$ and $\mathbf{\tilde{g}}$ to be small, avoiding severe deviation with each other.

The advantages that can be considered from each augmentation technique are: 1) local, global and adjacent views with the resized crop, 2) mirrored inputs with the horizontal flip, 3) color distortions with color jitter, 4) color independent facial representations with grayscale, and 5) noise in image with Gaussian blur. As a result, the contents included in the face image are robustly compared between the original and the transformed one in a self-similarity learning way to find the discriminative facial image representations.

Moving on to hash code learning, we present two loss functions to train $\mathcal{F}(\cdot)$ and $\mathcal{H}(\cdot)$. First, to cut down the gap between the discrete binary code and the real-valued hash code, we compute and minimize the squared Quantization loss as:

\begin{align}
\mathcal{L}_{sQ}(\mathbf{q})=\frac{1}{N_B\cdot K}\sum_{i=1}^{N_B}\sum_{j=1}^{K}\left(|1-q_{ij}^2|\right)
\label{equation:eqn3}
\end{align}

\noindent where $q_{ij}$ is the $j$-th element of $\mathbf{q}_i$, and $|\cdot|$ is the operation to return the absolute value. This term aims to minimize the quantization error and contributes to approximate the discrete binary code $\mathbf{b}$ with $\mathbf{q}$. Following \cite{DDQH}, we compute the loss before non-linear activation to increase the convergence speed and accuracy.

\begin{figure}[!t]
\centering
\includegraphics[width=0.99\linewidth, trim = {1cm 0 1cm 0}]{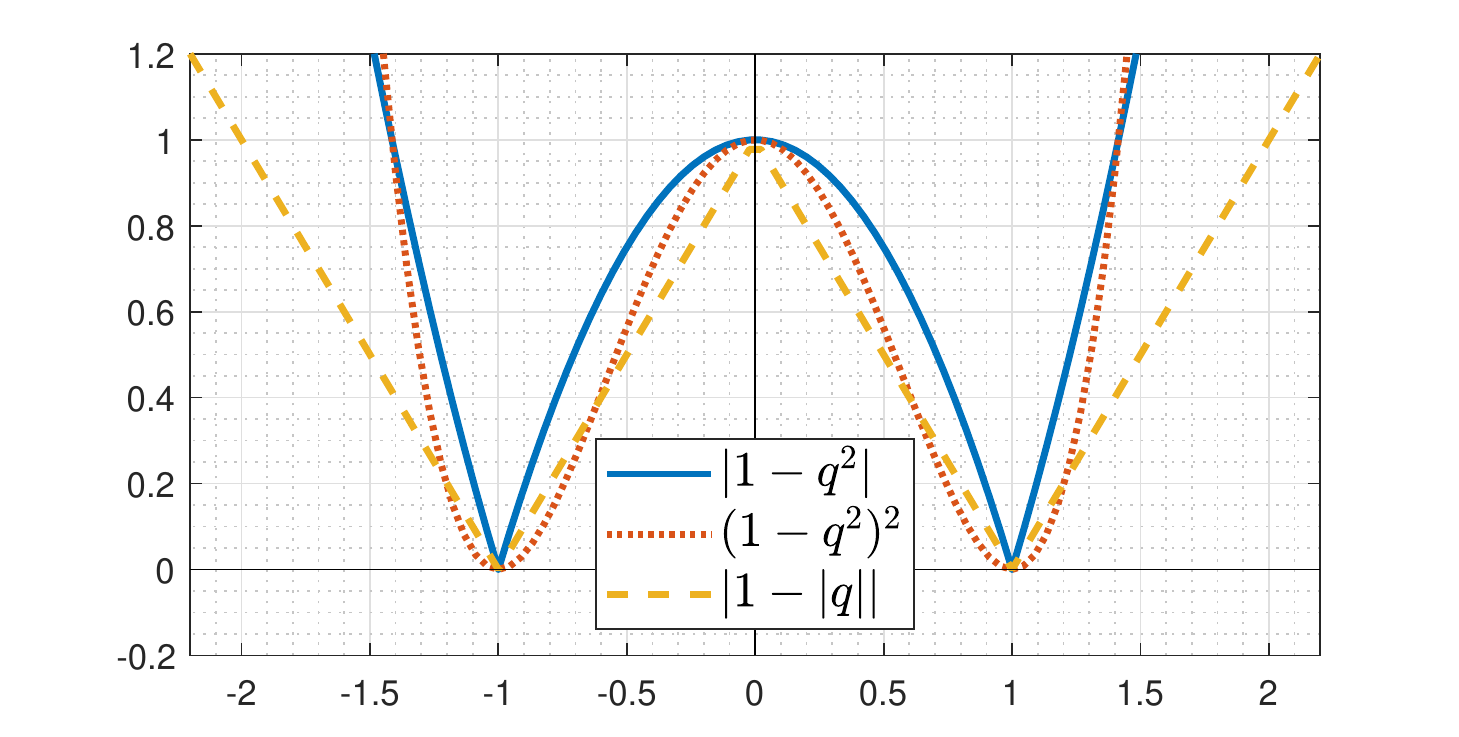}
\caption{An illustration of different quantization loss functions.}

\label{fig:Figure4}
\end{figure}

To compare with previously proposed quantization loss functions \cite{DDH, DCBH}, we plot each one, including ours in Figure \ref{fig:Figure4}. The double squared one ($(1-q^2)^2$, orange dotted line) has a convex shape near $-1$ and $1$, which makes it difficult to be strictly approximated, and the double absolute one ($|1-|q||$, yellow dashed line) generates relatively small error values, making a weak contribution. Therefore, we utilize the loss function (blue line) that can complement each other's shortcomings by combining the two.

Second, to make the binary code discriminative with the identity labels, we estimate the class probability on $\mathbf{h}$, which is the relaxed version of $\mathbf{b}$ as:

\begin{align}
\mathcal{L}_{cls}(\mathbf{h},\mathbf{y})=\frac{1}{N_B}\sum_{i=1}^{N_B}\mathcal{L}_{CE}(\text{FC}_{cls}(\mathbf{h}_i),\mathbf{y}_i)
\label{equation:eqn4}
\end{align}

\noindent where $\text{FC}_{cls}$ is a fully connected layer that outputs the probability prediction for all identities. In summary, high retrieval accuracy is achieved by taking into account the self and pairwise-similarity in latent space, as well as the general classification performance of the hash codes generated from the original and transformed images.

From the Equations~\ref{equation:eqn1} to~\ref{equation:eqn4}, the total training objective function $\mathcal{L}_{T}$ for the number of $N_B$ randomly sampled training samples can be formulated as:

\begin{align}
\mathcal{L}_{T}(\mathcal{X}_{N_B})=\mathcal{L}_{SP}+\lambda_{1}\mathcal{L}_{reg}+\lambda_{2}\mathcal{L}_{sQ}+\mathcal{L}_{cls}.
\label{equation:Eqn5}
\end{align}

\noindent where $\mathcal{X}_{N_B}=\{(x_1, \mathbf{y}_1),...,(x_{N_B}, \mathbf{y}_{N_B})\}$ and $\lambda_1$ and $\lambda_2$ are the hyperparameters that balance the contribution of each loss function. Note that, the entire SGH framework is trained with $\mathcal{L}_{T}$ in an end-to-end manner without excluding the contribution of any loss function.

We report detailed information on how SGH is trained for every iteration with a learning rate of $\gamma$ in Algorithm \ref{algorithm1}. Notably, SGH takes batch-axis concatenated inputs (notated as $[\cdot]_B$), therefore, actual batch size for training is $2 \times N_B$. Unlike other pairwise-similarity learning strategies such as Triplet \cite{FaceNet}, SGH does not require any complicated batch configuration process.

\begin{algorithm}[!ht]
\caption{SGH's main learning algorithm.}
\begin{algorithmic}[1]
\label{algorithm1}

\REQUIRE Trainable parameters : $\theta_{\mathcal{F}}, \theta_{\mathcal{G}}, \theta_{\mathcal{H}}$, batch size $N_B$
\FOR{sampled mini-batch $\{x_i, \mathbf{y}_i\}_{i=1}^{N_B}$}
\FOR{$i$ in 1...$N_B$}
\STATE draw an augmentation function $t_i\sim\mathcal{T}$
\STATE $\tilde{x}_i$ = $t_i(x_i)$
\STATE $[\mathbf{f}_i, \mathbf{\tilde{f}}_i]_B$ = $\mathcal{F}([x_i, \tilde{x}_i]_B)$
\STATE $[\mathbf{g}_i, \mathbf{\tilde{g}}_i]_B$ = $\mathcal{G}([\mathbf{f}_i, \mathbf{\tilde{f}}_i]_B)$
\STATE $[\mathbf{q}_i, \mathbf{\tilde{q}}_i]_B$ = $\mathcal{H}([\mathbf{f}_i, \mathbf{\tilde{f}}_i]_B)$
\STATE $[\mathbf{h}_i, \mathbf{\tilde{h}}_i]_B$ = $\text{tanh}([\mathbf{q}_i, \mathbf{\tilde{q}}_i]_B)$
\ENDFOR
\STATE $\ell_{SP} \leftarrow {\mathcal{L}_{SP}}$ with $\{\mathbf{g}_i, \mathbf{\tilde{g}}_i, \mathbf{y}_i\}_{i=1}^{N_B}$
\STATE $\ell_{2} \leftarrow {\mathcal{L}_{2}}$ with $\{\mathbf{g}_i, \mathbf{\tilde{g}}_i\}_{i=1}^{N_B}$
\STATE $\ell_{sQ} \leftarrow {\mathcal{L}_{sQ}}$ with $\{[\mathbf{q}_i, \mathbf{\tilde{q}}_i]_B\}_{i=1}^{N_B}$
\STATE $\ell_{cls} \leftarrow {\mathcal{L}_{cls}}$ with $\{[\mathbf{h}_i, \mathbf{\tilde{h}}_i]_B, [\mathbf{y}_i, \mathbf{y}_i]_B\}_{i=1}^{N_B}$
\STATE $\theta_\mathcal{F} \leftarrow{\theta_\mathcal{F} - \gamma \left(\frac{\partial \ell_{SP}}{\partial \theta_\mathcal{F}}+\frac{\partial \ell_{2}}{\partial \theta_\mathcal{F}}+\frac{\partial \ell_{sQ}}{\partial \theta_\mathcal{F}}+\frac{\partial \ell_{cls}}{\partial \theta_\mathcal{F}} \right)}$
\STATE $\theta_\mathcal{G} \leftarrow{\theta_\mathcal{G} - \gamma \left(\frac{\partial \ell_{SP}}{\partial \theta_\mathcal{G}}+\frac{\partial \ell_{2}}{\partial \theta_\mathcal{G}}\right)}$
\STATE $\theta_\mathcal{H} \leftarrow{\theta_\mathcal{H} - \gamma \left(\frac{\partial \ell_{sQ}}{\partial \theta_\mathcal{H}}+\frac{\partial \ell_{cls}}{\partial \theta_\mathcal{H}} \right)}$
\ENDFOR
\ENSURE Updated $\theta_\mathcal{F}, \theta_\mathcal{G}, \theta_\mathcal{H}$

\end{algorithmic}
\end{algorithm}\textbf{}

\section{Experiments}
\label{sec:4}

\begin{table}[!ht]
\centering
\caption{Detailed dataset configurations. `Ids' and `Resol'. are abbreviations for identities and resolution, respectively.}
\begin{adjustbox}{width=0.48\textwidth}
\begin{tabular}{c|c|c|c|c}
\toprule 
Dataset & \# Retrieval & \# Test & \#Ids & Resol. \\
\midrule
\midrule
\multicolumn{5}{c}{closed-set Protocol} \\
\midrule 
YouTube Faces & 63,800 & 7,975 & 1,595 & $32\times32$\\
FaceScrub & 67,177 & 2,650 & 530 & $32\times32$\\
VGGFace2-Test & 75,296 & 2,500 & 500 & $96\times96$\\
\midrule
\multicolumn{5}{c}{Open-set Protocol} \\
\midrule 
VGGFace2-Train & 17,940 & 500 & 100 & $96\times96$\\
\bottomrule 
\end{tabular}
\end{adjustbox}
\label{table:Table2}
\end{table}

\begin{table*}[!t]
\centering
\caption{mAP scores of different hashing approaches on small face image datasets.}
\begin{adjustbox}{width=0.78\textwidth}
\begin{tabular}{clcccccccc}
\toprule
\multicolumn{2}{c}{\multirow{2}{*}{Method}} & \multicolumn{4}{c}{YouTube Faces}                                    & \multicolumn{4}{c}{FaceScrub}                                        \\
\multicolumn{2}{c}{}                        & 12-bits          & 24-bits          & 36-bits          & 48-bits          & 12-bits          & 24-bits          & 36-bits          & 48-bits          \\ \midrule\midrule
\multicolumn{2}{c}{ITQ \cite{ITQ} + CNN}                     & 0.0248          & 0.1900         & 0.3420         & 0.4394          & 0.0186          & 0.0352          & 0.0504          & 0.0667          \\ \midrule
\multicolumn{2}{c}{SH \cite{SH} + CNN}                    & 0.0154          & 0.0851          & 0.1603          & 0.2421          & 0.0036         & 0.0081          & 0.0114          & 0.0145          \\ \midrule
\multicolumn{2}{c}{KSH \cite{KSH} + CNN}                     & 0.0481          & 0.2663          & 0.4167          & 0.5047          & 0.0230          & 0.0348          & 0.0767          & 0.1026          \\ \midrule
\multicolumn{2}{c}{SDH \cite{SDH} + CNN}                    & 0.5474          & 0.7676          & 0.8100          & 0.8331          & 0.1281          & 0.2388          & 0.2934          & 0.3291          \\ \midrule
\multicolumn{2}{c}{DSH \cite{DSH}}                     & 0.5034         & 0.6011          & 0.7132          & 0.7354          & 0.5341          & 0.5955          & 0.6112          & 0.6234          \\ \midrule
\multicolumn{2}{c}{DHCQ \cite{DHCQ}}                     & 0.8108          & 0.8892          & 0.9122          & 0.9258          & 0.4986          & 0.5834          & 0.6215          & 0.6387          \\ \midrule
\multicolumn{2}{c}{DDH \cite{DDH}}                     & 0.8808          & 0.9212          & 0.9340          & 0.9412          & 0.5985          & 0.6121          & 0.6445          & 0.6765          \\ \midrule
\multicolumn{2}{c}{DDH-Deep}                & 0.9322          & 0.9455          & 0.9580          & 0.9718          &  0.6215          & 0.6479          & 0.6883          & 0.6983          \\ \midrule
\multicolumn{2}{c}{DDQH \cite{DDQH}}                     & 0.9002          & 0.9553          & 0.9684          & 0.9820          & 0.6452          & 0.6824          & 0.7120          & 0.7355          \\ \midrule
\multicolumn{2}{c}{DDQH-Deep}               & 0.9580          & 0.9782          & 0.9790          & 0.9834           & 0.6618          & 0.7122          & 0.7654          & 0.7798          \\ \midrule
\multicolumn{2}{c}{DCBH \cite{DCBH}}                & 0.9753          & 0.9899          & 0.9914          & 0.9922          &  0.7182          & 0.7317          & 0.7696          & 0.7862          \\ \midrule
\multicolumn{2}{c}{DAGH \cite{DAGH}}               & 0.9744          & 0.9926          & 0.9938          & 0.9946          & 0.7284          & 0.7919          & 0.8172          & 0.8204          \\ \midrule
\multicolumn{2}{c}{\textbf{SGH}}                    & \textbf{0.9902} & \textbf{0.9933} & \textbf{0.9955} & \textbf{0.9966} &  \textbf{0.8970}   & \textbf{0.9219}   & \textbf{0.9319} & \textbf{0.9345}  \\ \bottomrule
\end{tabular}
\end{adjustbox}
\label{table:Table3}
\end{table*}

\subsection{Datasets}

Following the closed-set protocol proposed in \cite{DHCQ, DDH, DDQH, DCBH} where the retrieval database includes training samples, we conduct experiments on two well-organized face image retrieval datasets and a high-resolution face image dataset that we constructed. 

\textit{YouTube Faces} dataset (Y.T.F) \cite{YTF} is composed of video frames containing faces, which is designed to study unconstrained face recognition problems. There are 1,595 different identities to classify, where 40 images for each person are randomly selected to be the training data, and five images per person are utilized for the test data. A total of 63,800 images are used to build a training set and a retrieval database, and 7,975 images are used as a query dataset for the experiment. 

\textit{FaceScrub} dataset (F.S) \cite{FS} consists of 106,863 face images which are collected from the Internet. There are 530 celebrity classes to distinguish with about 200 images per person. We choose five images for each person to utilize a total of 2,650 images in the experiment, and the remaining images are used for training and constructing a retrieval database. For both datasets above, the size of the image is fixed at $32 \times 32$.

In addition, we configure a new higher resolution face image retrieval dataset with \textit{VGGFace2} (V.F2) \cite{VGGFace2}, which contains over 3.3 million images of more than 9,000 identities in the training set, and about 0.15 million images of 500 identities in the test set. We apply a fast and accurate deep learning-based face detection algorithm DSFD \cite{DSFD} to the images in the \textit{test set} of V.F2 to extract strict face images. We select the images whose width and height do not exceed 1,024 for detection and utilize the output face images that have confidence scores over 0.99 to build a dataset. Also, all extracted images have been resized to $96 \times 96$, and images that do not exceed 96 in either width and height are discarded during this process. As a result, we could collect a total of 75,296 face images of 500 different identities. Similar to existing dataset protocols, we split the dataset into two as 72,796 images for the training set, and the rest 2,500 images for the test (5 images per identity).

Following the identity-disjoint (open-set) protocol proposed in \cite{DAGH}, we randomly sample 100 identities from the \textit{training set} of V.F2 with a total of 17,940 images, which do not participate in the training process. The same face detection and resizing method are applied to conduct retrieval with the model trained from V.F2 closed-set protocol. Five images per identity are randomly selected for testing, and the remaining images are utilized to build a retrieval database. The exact numbers are reported in Table \ref{table:Table2}.

\subsection{Evaluation setups}

To evaluate quality of the face image retrieval outputs, we employ four metrics: 1) mean average precision (\textbf{mAP}), 2) precision within Hamming distance 2 (\textbf{P@H$\leq$2}) for different bit-lengths, 3) precision-recall curves (\textbf{PR curves}), and 4) precision with respect to top-$M$ returned image (\textbf{P@Top-$M$}). In terms of computing mAP scores, we select top 50 images from the retrieval ranked-list results. The retrieval accuracy is estimated based on whether the returned images and the query image have the same identity label or not. We set the length of binary codes as 12, 24, 36, and 48 to examine the performance according to the number of bits.

\subsection{Implementation Details}

For non-deep learning hashing approaches; ITQ \cite{ITQ}, SH \cite{SH}, KSH \cite{KSH} and SDH \cite{SDH}, we utilize public available source codes. The notation "+CNN" in Table \ref{table:Table3} denotes taking deep feature inputs which are originating from DHCQ \cite{DHCQ} model, and we utilize the publicly open source codes to conduct face image retrieval experiments. In the case of deep learning-based hashing methods; DSH \cite{DSH}, DHCQ \cite{DHCQ}, DDH \cite{DDH}, DDQH \cite{DDQH}, DCBH \cite{DCBH}, DAGH \cite{DAGH} and our SGH, we implement them with PyTorch framework and train with NVIDIA GeForce RTX 2080 Ti.

When it comes to deep network training, we employ Adam optimizer \cite{Adam} with the initial learning rate of 0.001, which decays by 0.9 times for every 50-th epoch. The balancing hyperparameter $\lambda_1$ and $\lambda_2$ are set to 0.0002 and 0.05, respectively, as default. The dimensionality of $\mathbf{f}$ and $\mathbf{g}$ are fixed as $D_{\mathbf{f}}=512$ and $D_{\mathbf{g}}=128$. We fix the batch size $N_B$ as 256, however, since we apply randomly sampled data augmentation techniques on every image to the obtained transformed images, the actual batch size that passes through the deep network is doubled to 512.

\begin{figure*}[!t]
\centering
\subfigure[P@H$\leq$2 for different bit-lengths]{
\includegraphics[width=0.31\linewidth]{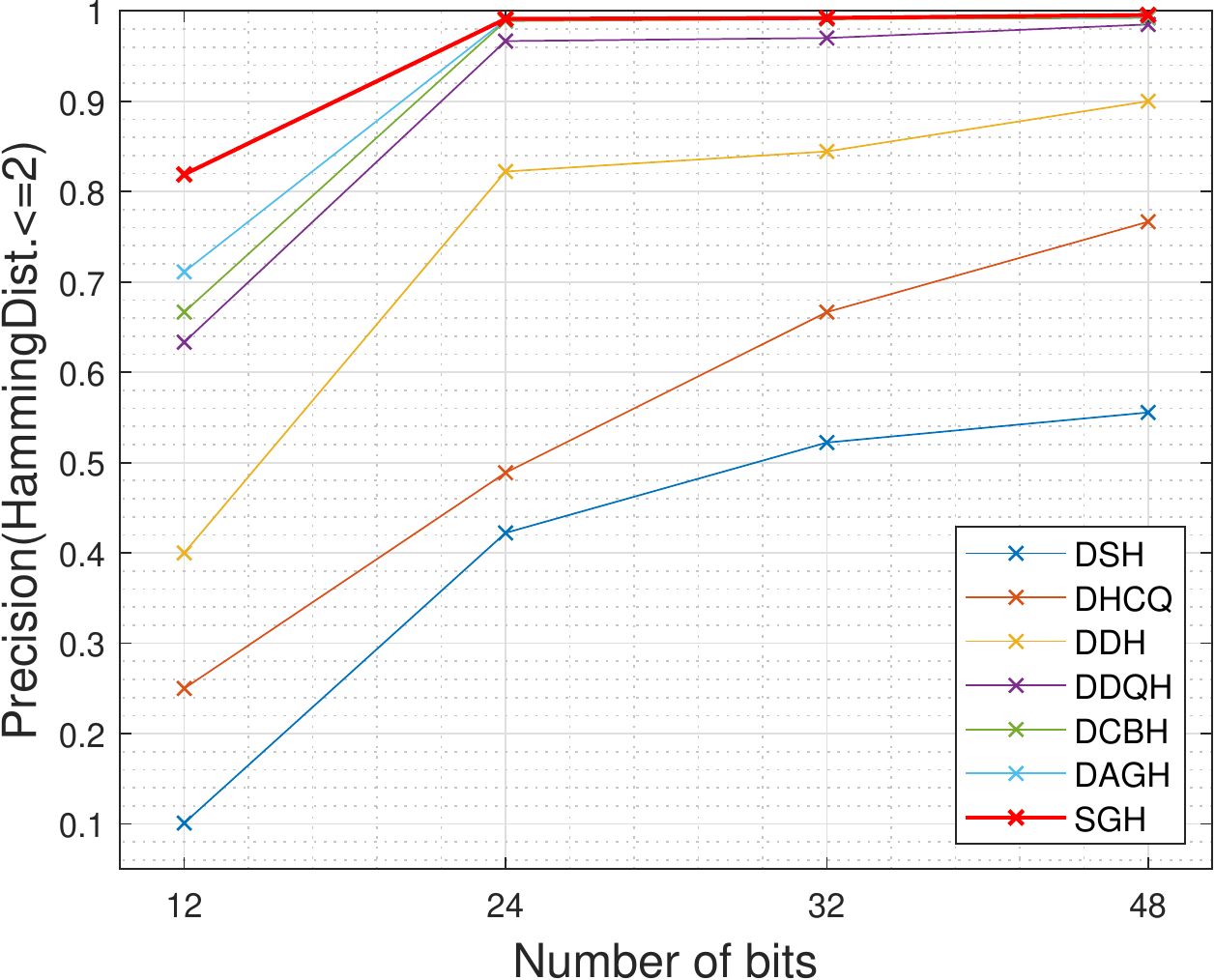}
}
\subfigure[PR curves @48-bits]{
\includegraphics[width=0.31\linewidth]{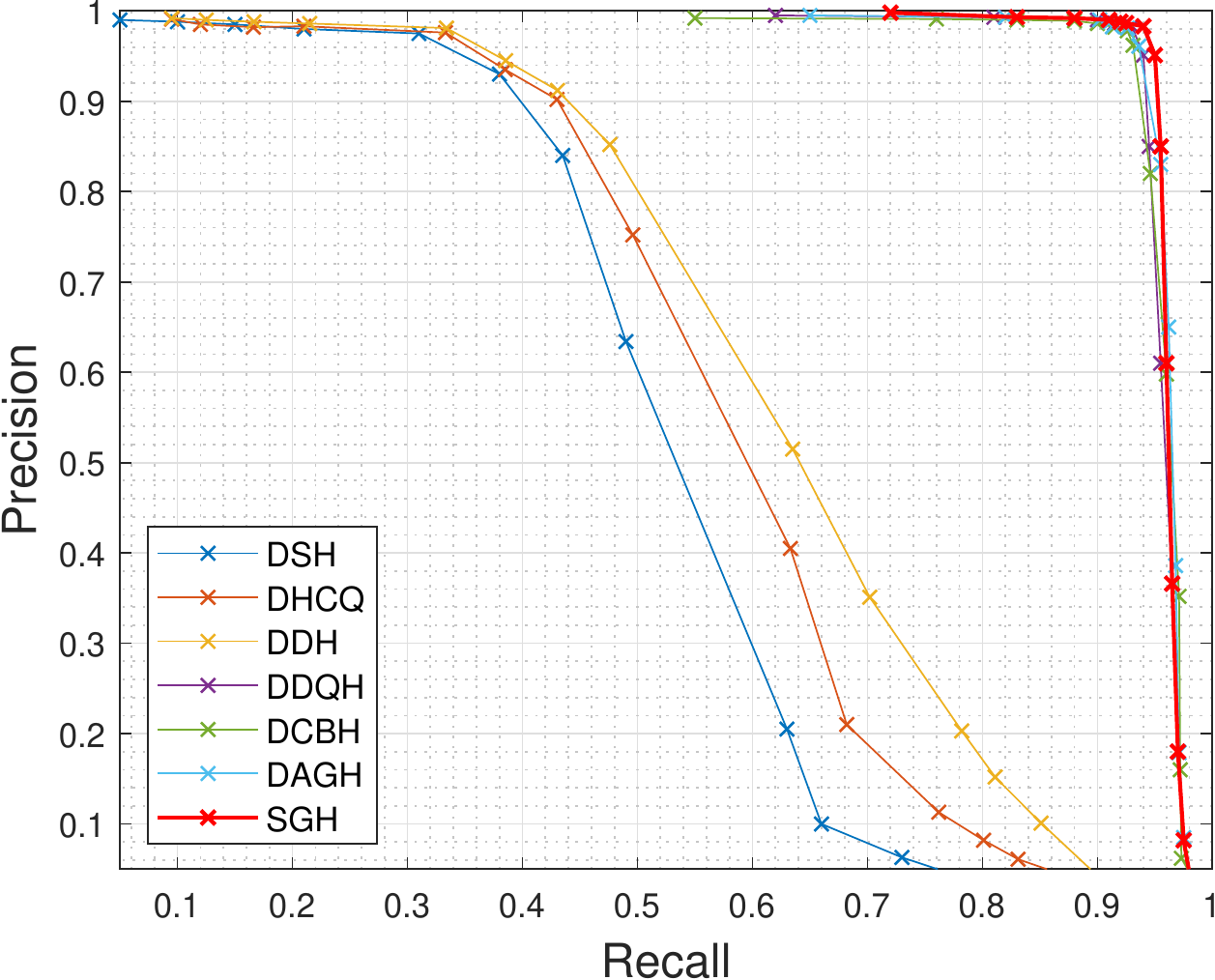}
}
\subfigure[P@Top-$M$ @48-bits]{
\includegraphics[width=0.31\linewidth]{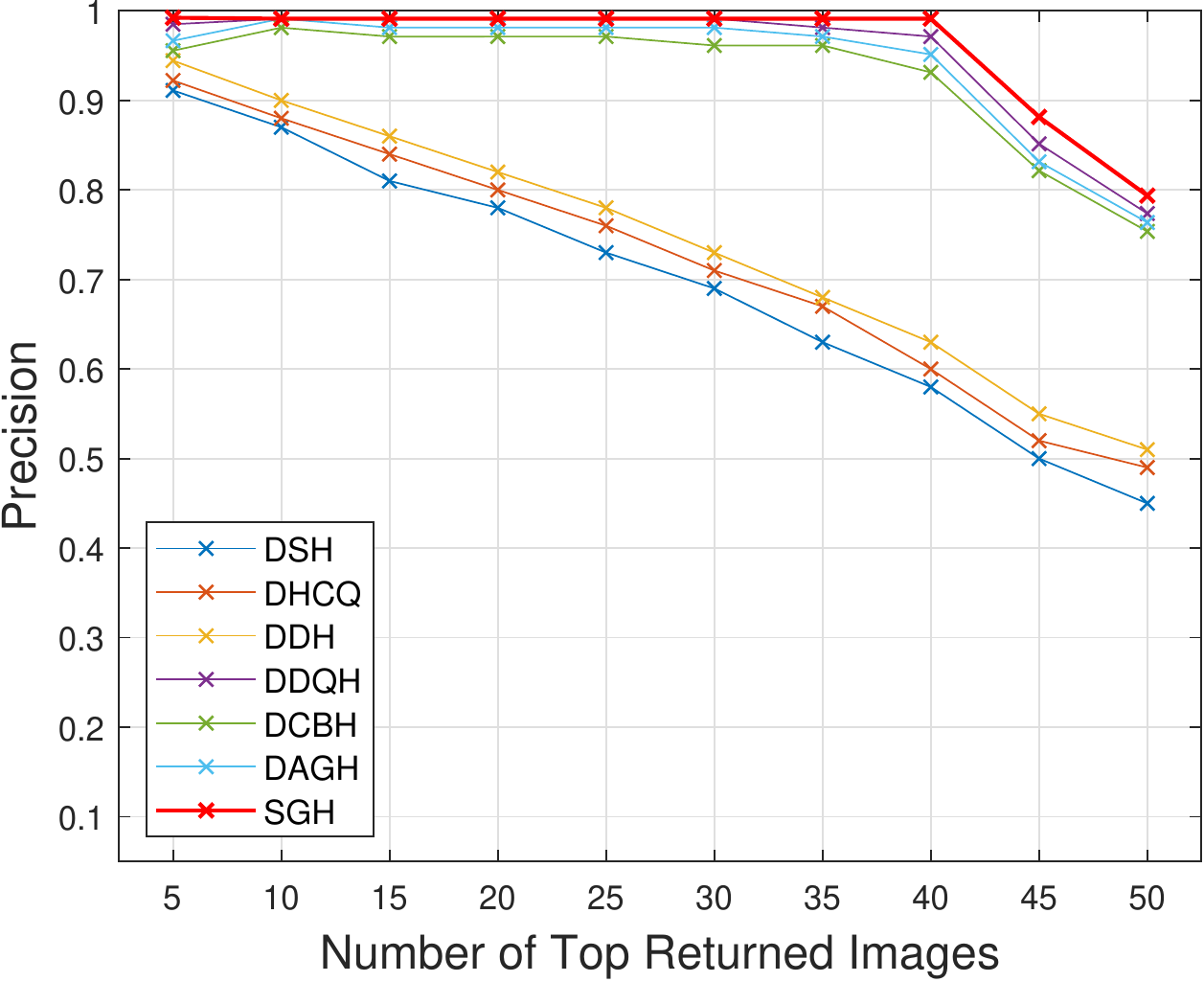}
}
\caption{Experimental results under closed-set protocol with YouTube Faces dataset.}

\label{fig:Figure5}
\end{figure*}

\begin{figure*}[!t]
\centering
\subfigure[P@H$\leq$2 for different bit-lengths]{
\includegraphics[width=0.31\linewidth]{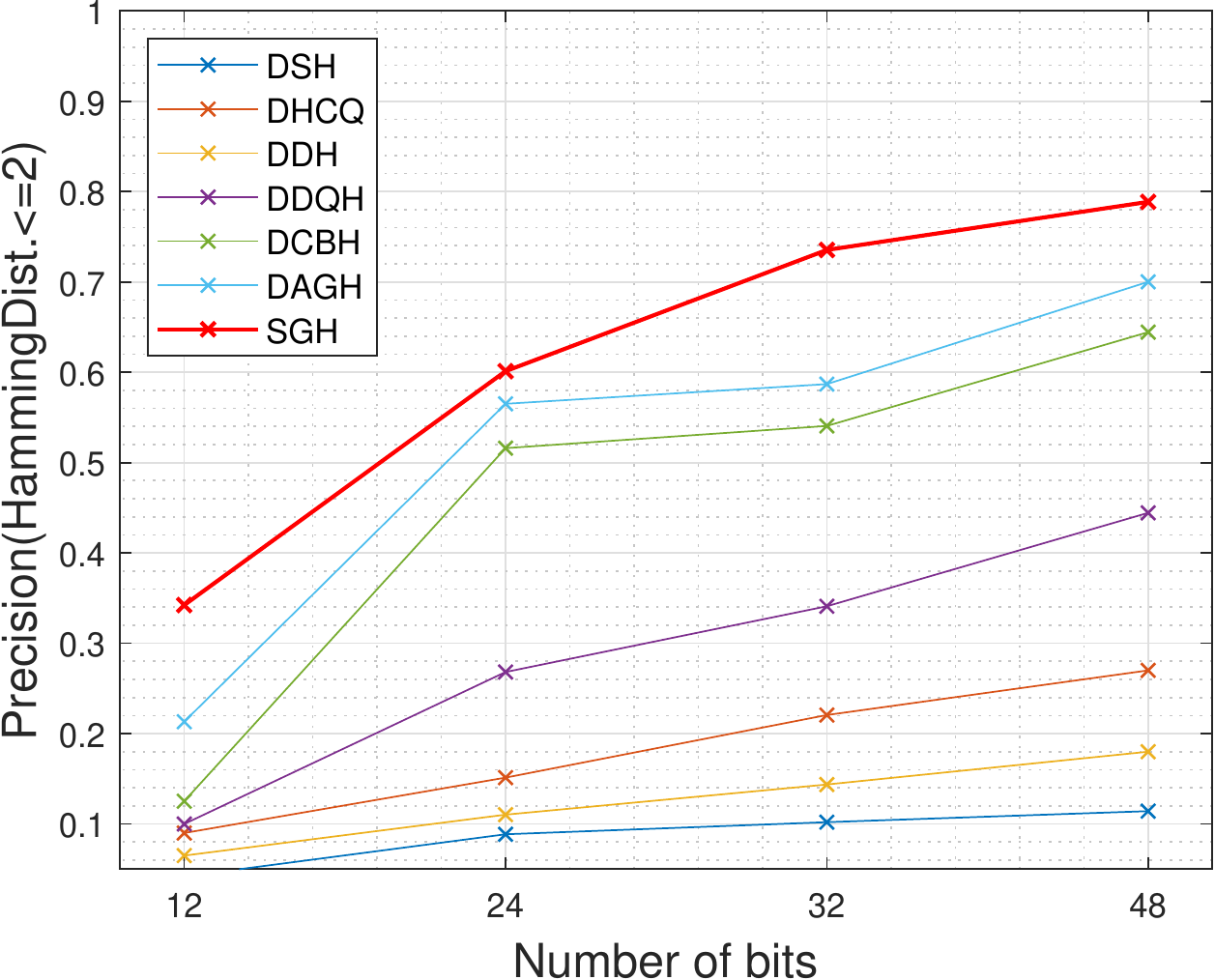}
}
\subfigure[PR curves @48-bits]{
\includegraphics[width=0.31\linewidth]{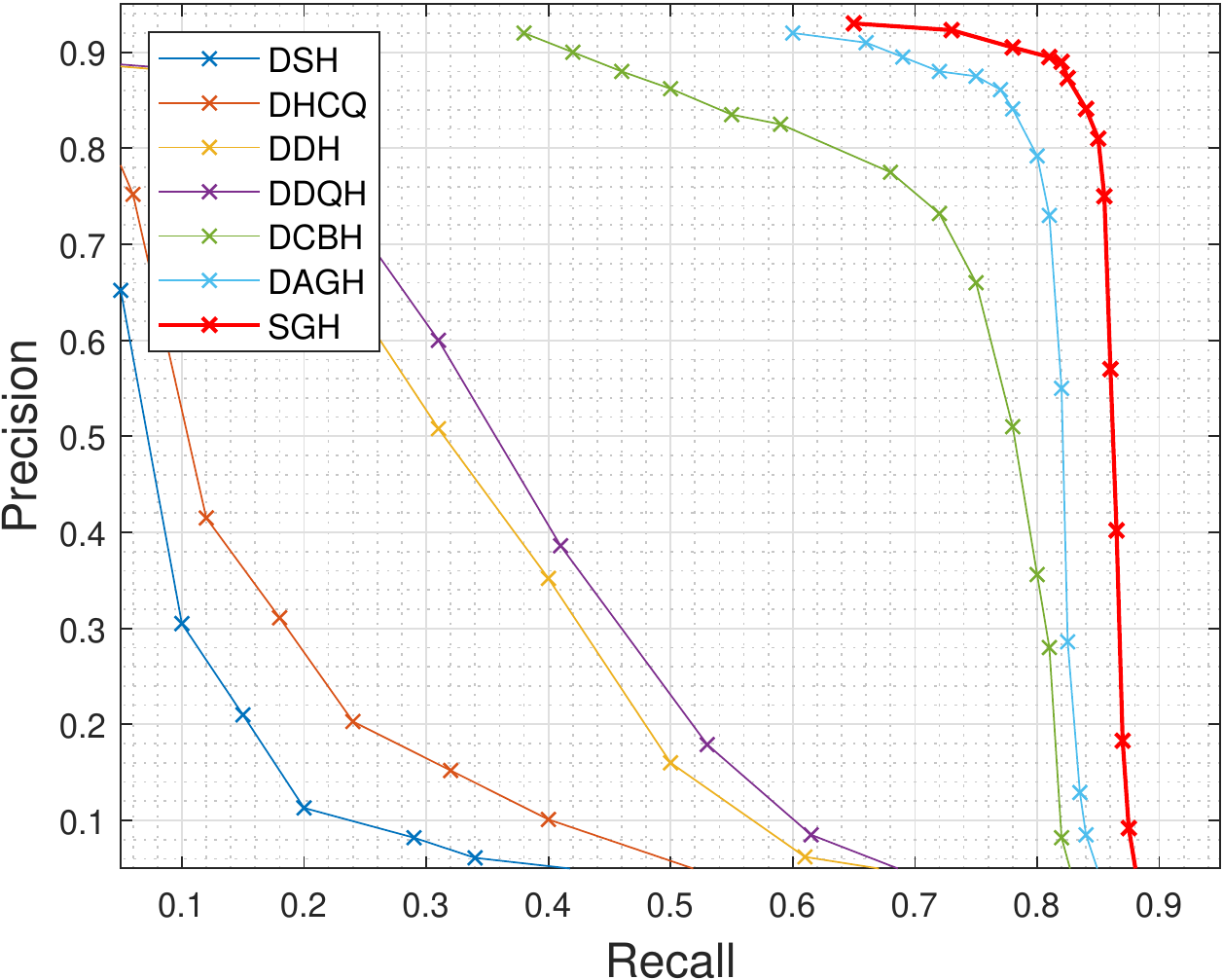}
}
\subfigure[P@Top-$M$ @48-bits]{
\includegraphics[width=0.31\linewidth]{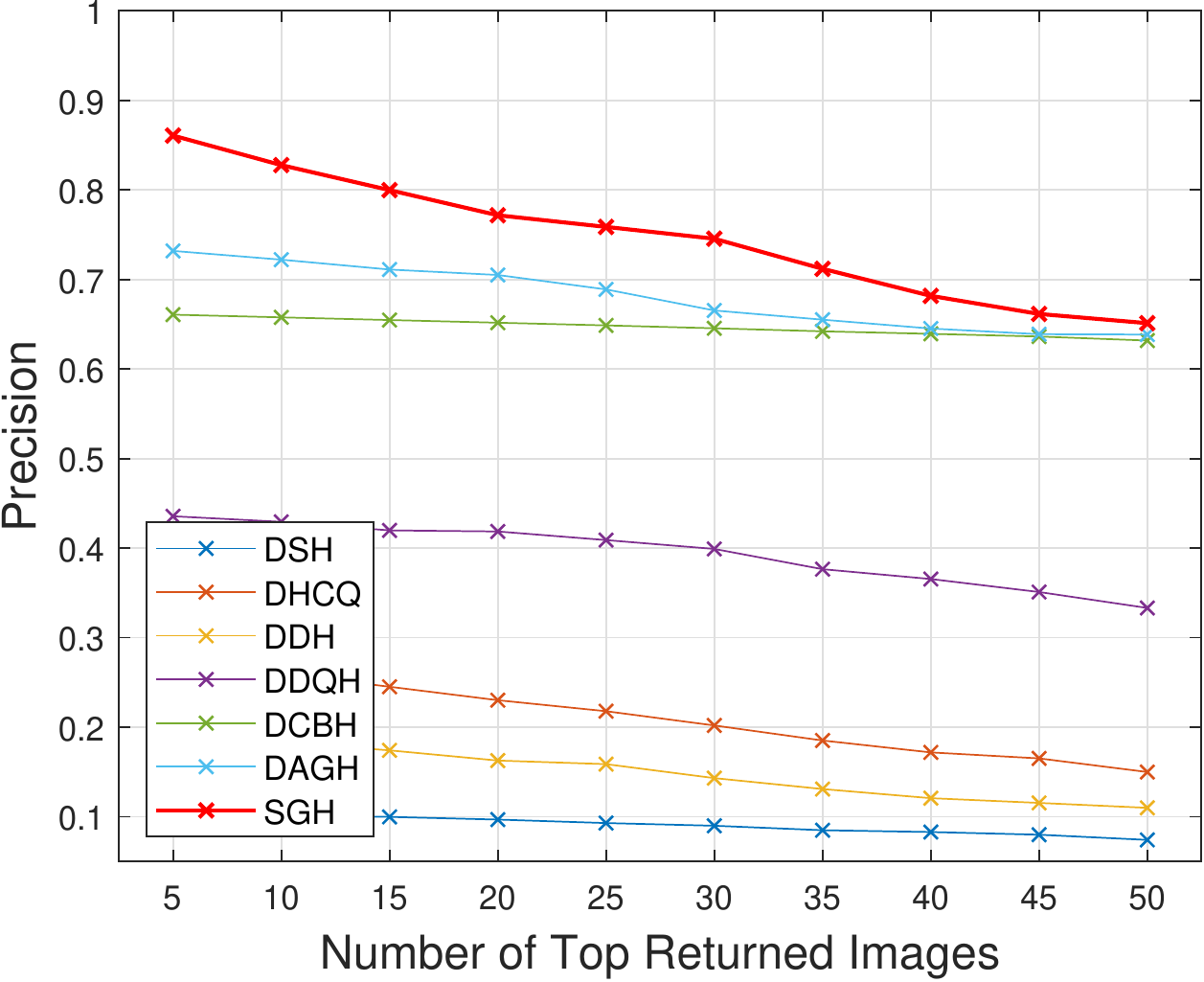}
}
\caption{Experimental results under closed-set protocol with FaceScrub dataset.}

\label{fig:Figure6}
\end{figure*}

\subsection{Results on Small Face Images}

As shown in Table \ref{table:Table1}, we calculate the mAP scores on small face-image datasets; Y.T.F and F.S for both non-deep learning-based and deep learning-based hashing approach over four types of bit-lengths. Deep hashing methods generally outperform non-deep hashing ones because expensive and effective identity annotations are fully utilized during deep model training, which helps deep hash codes to be discriminative according to the labels. Noting the case of SDH that employs supervised signals, it also shows promising outcomes even though deep learning is not exploited.

To verify that our similarity guided training scheme and squared Quantization loss are effective, we conduct experiments with DDH-Deep and DDQH-Deep, which share the same network architecture with SGH, noting that network architecture is the only factor that contributes to improvement. In addition, since DCBH and DAGH have claimed their contributions to the network architecture, we do not add any modification to them. Following the experimental results shown in Table \ref{table:Table2} and Figure \ref{fig:Figure5}, \ref{fig:Figure6}, our SGH can achieve the best performance over all bit-lengths, and accurately retrieves relevant images more in the top ranks.

\begin{table}[!t]
\centering
\caption{mAP scores of different hashing approaches on large face image dataset.}
\begin{adjustbox}{width=0.45\textwidth}
\begin{tabular}{ccccc}
\toprule 
\multicolumn{5}{c}{VGGFace2} \\
Method & 12-bits & 24-bits & 36-bits & 48-bits \\ 
\midrule
\midrule
\multicolumn{5}{c}{closed-set Protocol} \\
\midrule
DDH-Deep &0.3385  &0.3815  &0.4120 & 0.4424\\
DDQH-Deep &0.3507  &0.3934  &0.4324 & 0.4658\\
DCBH &0.6612  &0.6882  &0.7084 & 0.7254\\
DAGH &0.6576  &0.6998  &0.7273 & 0.7556\\
\textbf{SGH} &\textbf{0.8521}   & \textbf{0.8886}   & \textbf{0.9046} & \textbf{0.9174}\\
\midrule
\multicolumn{5}{c}{Open-set Protocol} \\
\midrule
DDH-Deep &0.2154  &0.2689  &0.3413 & 0.3995\\
DDQH-Deep &0.2193  &0.2872  &0.3736 & 0.4285\\
DCBH &0.2456  &0.3013  &0.3885 & 0.4313\\
DAGH &0.2334  &0.3126  &0.3906 & 0.4425\\
\textbf{SGH} &\textbf{0.3342}   & \textbf{0.4380}   & \textbf{0.4920} & \textbf{0.5594}\\
\bottomrule
\end{tabular}
\end{adjustbox}
\label{table:Table4}
\end{table}

\subsection{Results on Large Face Images}

The images included in Y.T.F and F.S are too small to represent detail facial characteristics. Therefore, we construct a new dataset with VGGFace2 where the resolution of all images is three times higher to compare the retrieval performance of SGH with the existing deep face image retrieval algorithms. As observed in Table \ref{table:Table4}, SGH exceeds all compared methods by large margins in both the closed-set and open-set protocols, even if the network architecture is the same (DDH-Deep, DDQH-Deep).

To be specific with the results, in the case of closed-set protocol, SGH demonstrates that generating better quality hash codes than the other deep face image hashing approaches, similar to the experimental outcomes observed on the small face image datasets. Especially for open-set protocol, SGH still significantly outperforms others by not overfitted to the training dataset but keep the generalization capacity within in the hashing head. In a nutshell, the proposed loss functions and training strategy of SGH make the deep network generate descriptive face hash codes from both small and large face images to conduct accurate retrieval.

\subsection{Ablation Study}

To investigate the effects of our proposed loss functions, we perform an ablation study on face retrieval benchmarks by measuring the mAP on 12-bit and 48-bit binary codes. Eliminating $\mathcal{L}_{cls}$ is ruled out because $\mathcal{H}$ cannot be trained without it, causing severe performance degradation. As shown in Table~\ref{table:Table5}, we can easily find that the baseline with all the training objectives achieves the best results on all datasets. Especially for without $\mathcal{L}_{SP}$, the result shows significant performance difference, which demonstrates that learning the self-similarity and the pairwise-similarity in the latent space has a huge influence on retrieval quality improvement.

\begin{figure}[!t]
\centering
\subfigure[@12-bits]{
\includegraphics[width=0.45\linewidth]{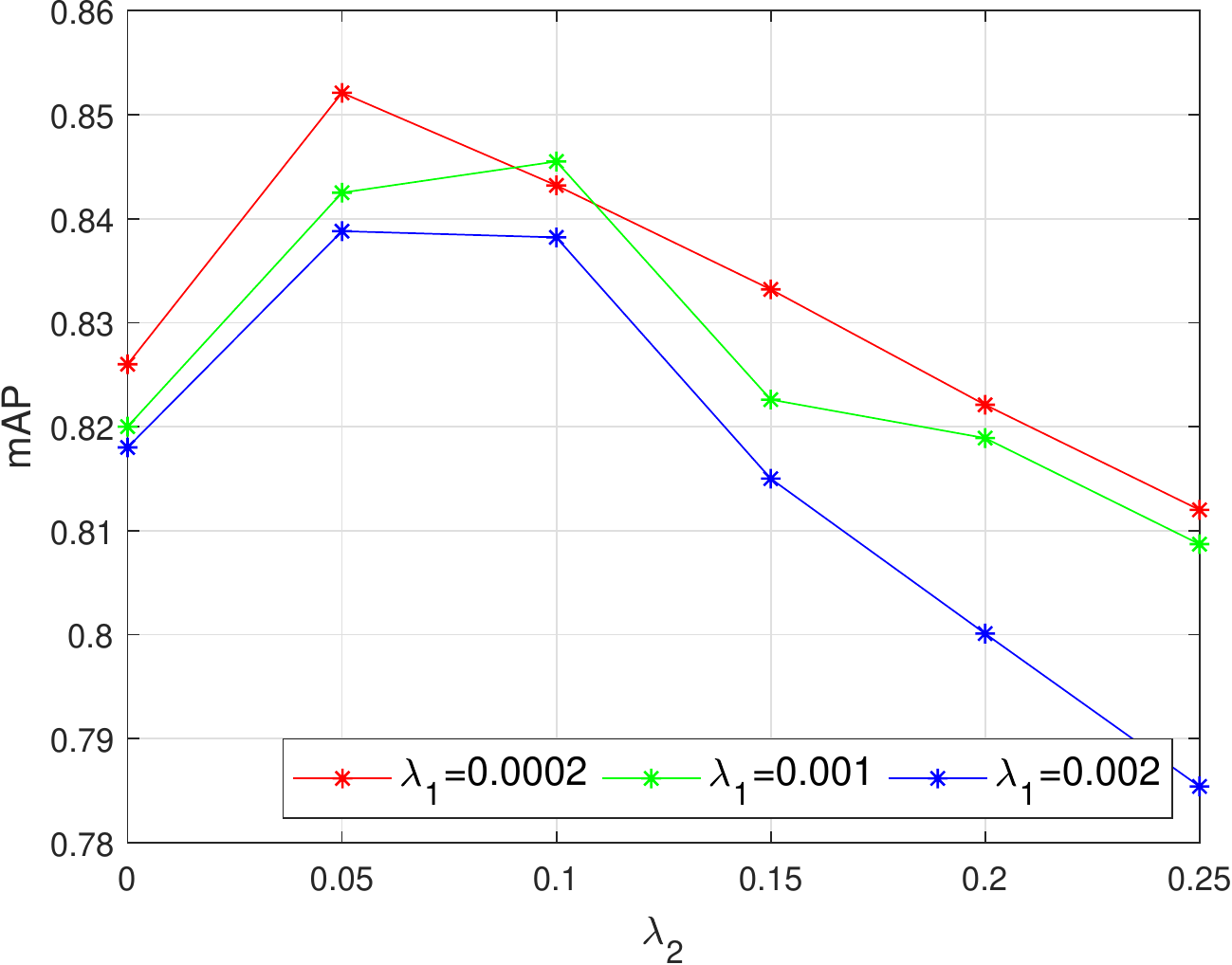}
}
\subfigure[@48-bits]{
\includegraphics[width=0.45\linewidth]{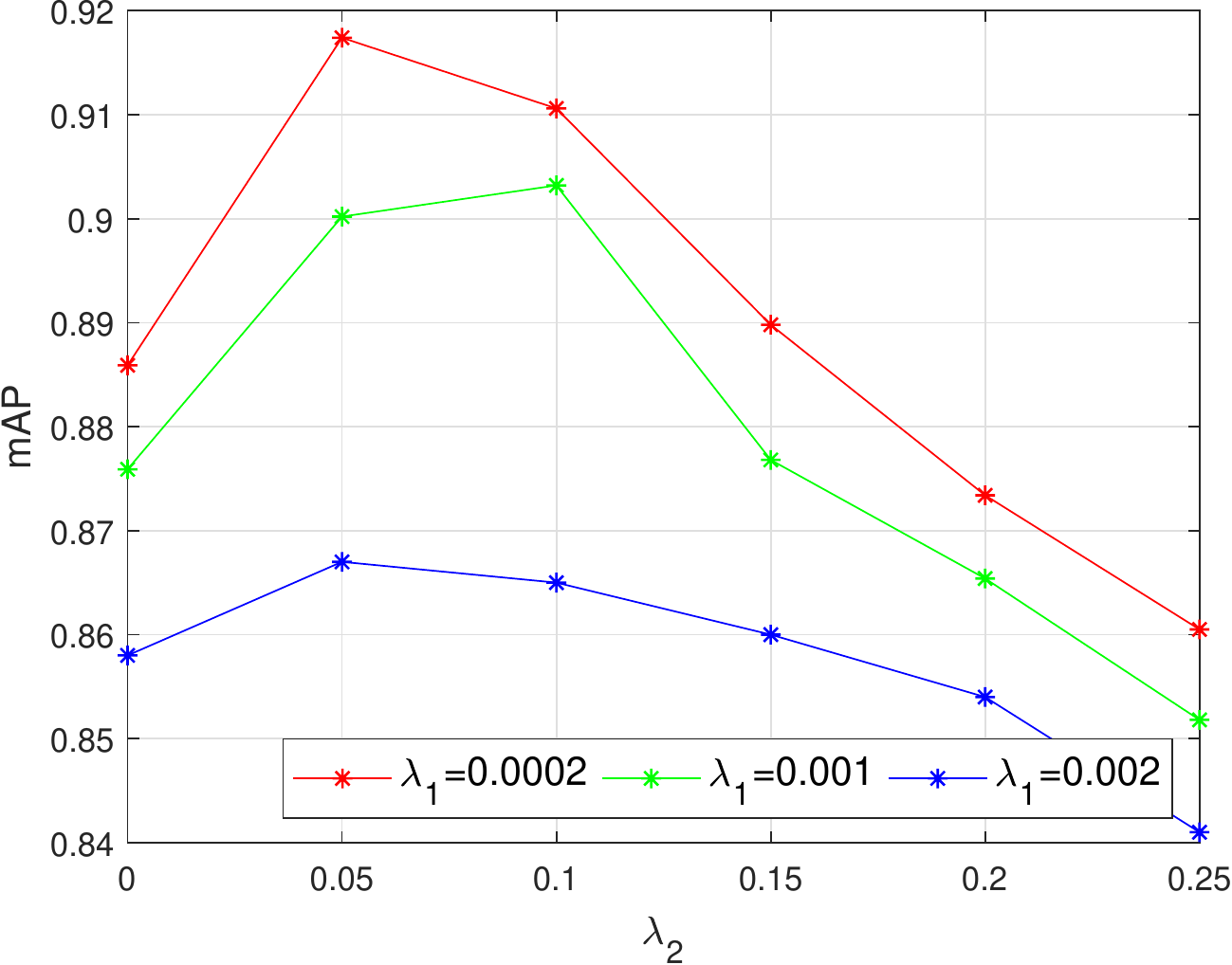}
}
\caption{The sensitivity investigation of $\lambda_1$ and $\lambda_2$ for VGGFace2 closed-set.}

\label{fig:Figure7}
\end{figure}

\begin{figure}[!t]
\centering
\subfigure[YouTube Faces]{
\includegraphics[width=0.99\linewidth]{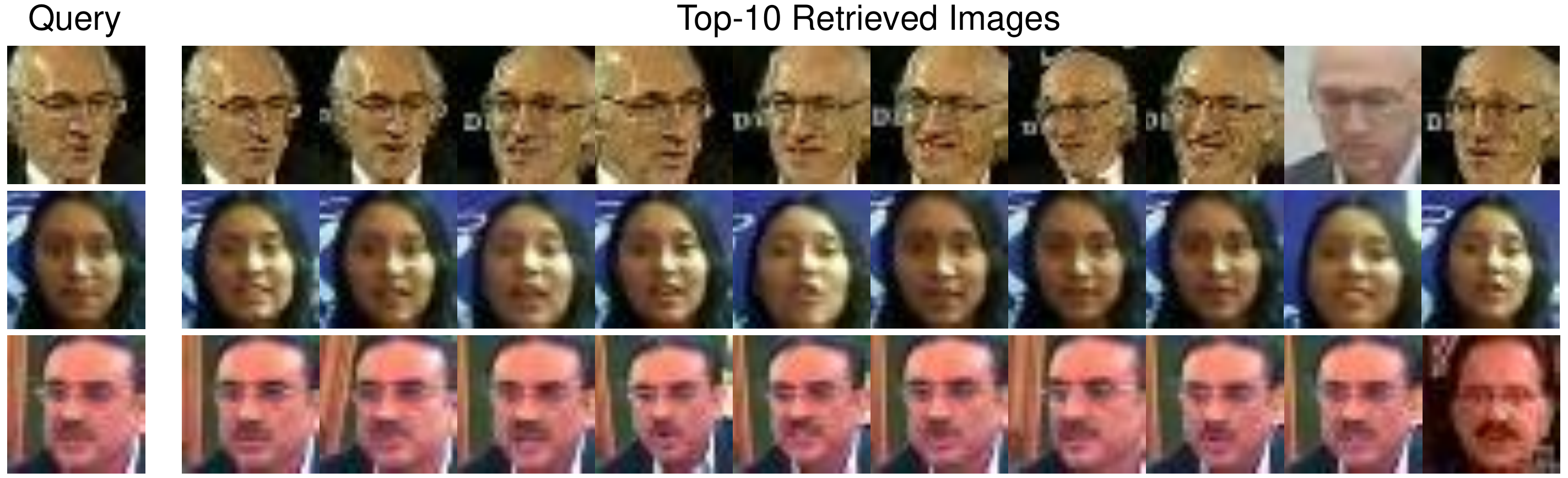}
}
\subfigure[FaceScrub]{
\includegraphics[width=0.99\linewidth]{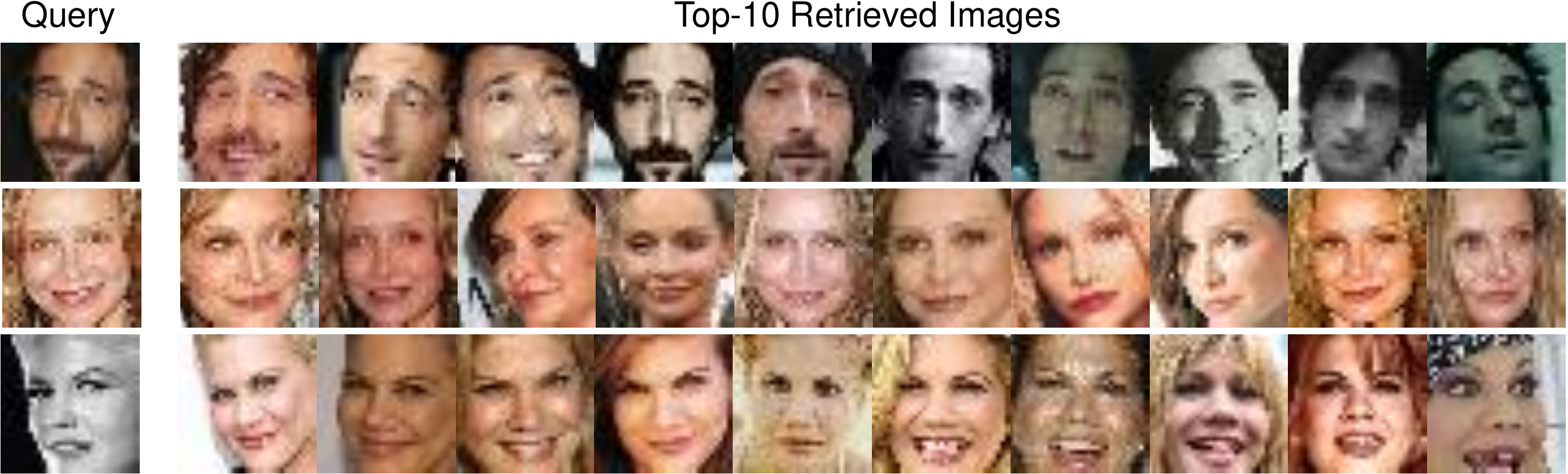}
}
\subfigure[VGGFace2 closed-set]{
\includegraphics[width=0.99\linewidth]{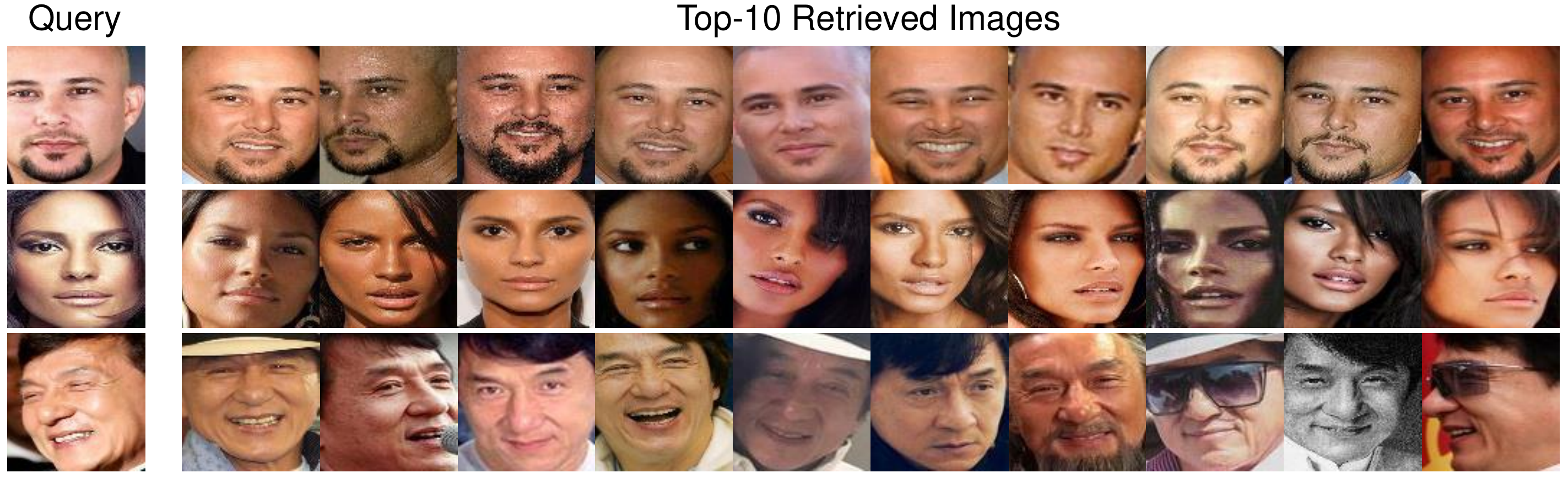}
}
\subfigure[VGGFace2 Open-set]{
\includegraphics[width=0.99\linewidth]{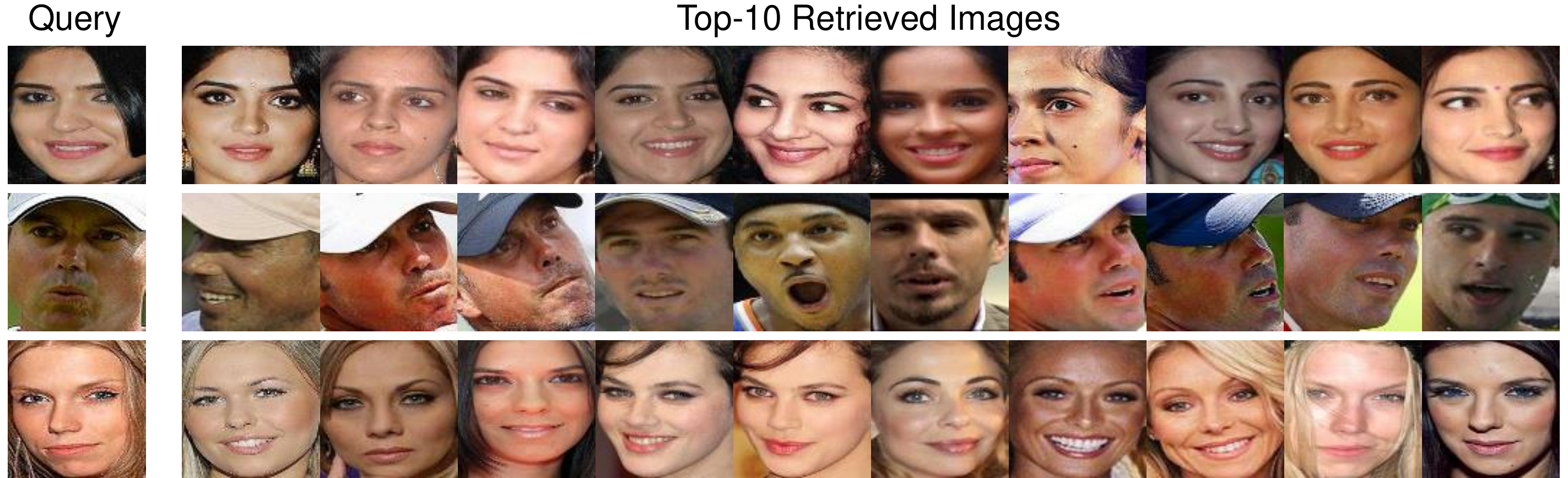}
}
\caption{Retrieval results on face image datasets @48-bits.}

\label{fig:Figure8}
\end{figure}

Experimental results for different options on hyperparameters $\lambda_1$ and $\lambda_2$ are detailed in Figure \ref{fig:Figure7}. As observed, if the $\lambda_1$ is too high, the image representations in the latent space are restricted, resulting in a decrease in retrieval performance. Similarly, if an excessively high value is set to $\lambda_2$, performance degradation may occur by limiting the representations in the hash space. Therefore, we elaborately adjust both hyperparameters to derive the optimal results. In addition, the actual retrieval results in Figure \ref{fig:Figure8} exhibits prominent performance of our SGH in returning relevant face images.

As reported in Table \ref{table:Table6}, we investigate the effect of each data augmentation technique used in SGH on retrieval performance, by excluding one by one. From the results, we can see that every transformation contributes to improving accuracy. Specifically, the resized crop contributes the most, and the Gaussian blur makes the least contribution in improving face image retrieval system.

\begin{table}[!t]
\caption{Ablation study results of our proposed SGH. `w/o' is an abbreviation for without.}
\centering
\begin{adjustbox}{width=0.49\textwidth}
\begin{tabular}{ccccc}
\toprule 
Dataset & \textbf{Baseline} & w/o $\mathcal{L}_{SP}$ & w/o $\mathcal{L}_{2}$ & w/o $\mathcal{L}_{sQ}$ \\ 
\midrule
\midrule
\multicolumn{5}{c}{12-bits} \\
\midrule
Y.T.F & \textbf{0.9902}  &0.9660  &0.9718 & 0.9824\\
F.S & \textbf{0.8970}  &0.7006  &0.8542 & 0.8662\\
V.F2-closed & \textbf{0.8521}  &0.6782  &0.8023 & 0.8181\\
V.F2-Open & \textbf{0.3342}  &0.2580  &0.3113 & 0.3180\\
\midrule
\multicolumn{5}{c}{48-bits} \\
\midrule
Y.T.F & \textbf{0.9966}  &0.9855  &0.9910 & 0.9906\\
F.S &\textbf{0.9345}  &0.7912  &0.9008 & 0.9125\\
V.F2-closed &\textbf{0.9174}  &0.7824  &0.8709 & 0.8816\\
V.F2-Open &\textbf{0.5594}  &0.4329  &0.5209 & 0.5335\\
\bottomrule
\end{tabular}
\end{adjustbox}
\label{table:Table5}
\end{table}

\begin{table}[!t]
\centering
\caption{mAP of different transformation compositions. `w/o' is an abbreviation for without.}
\begin{adjustbox}{width=0.49\textwidth}
\begin{tabular}{cccc}
\toprule 
Method & YouTube Faces & FaceScrub & VGGFace2\\
\midrule
\midrule
\multicolumn{4}{c}{@48-bits} \\
\midrule
w/o Resized Crop & 0.9896          & 0.8342         & 0.8098\\
\midrule
w/o Horizontal Flip & 0.9911          & 0.8867         & 0.8540\\
\midrule
w/o Color jitter & 0.9922          & 0.9059         & 0.8847\\
\midrule
w/o Grayscale & 0.9930          & 0.9213         & 0.8989\\
\midrule
w/o Gaussian Blur & 0.9933          & 0.9217         & 0.9006\\
\midrule
with all transforms & 0.9966          & 0.9345         & 0.9174\\
\bottomrule
\end{tabular}
\end{adjustbox}
\label{table:Table6}
\end{table}

\begin{figure}[!t]
\centering
\subfigure[t-SNE visualization of YouTube Faces dataset.]{
\includegraphics[width=0.99\linewidth]{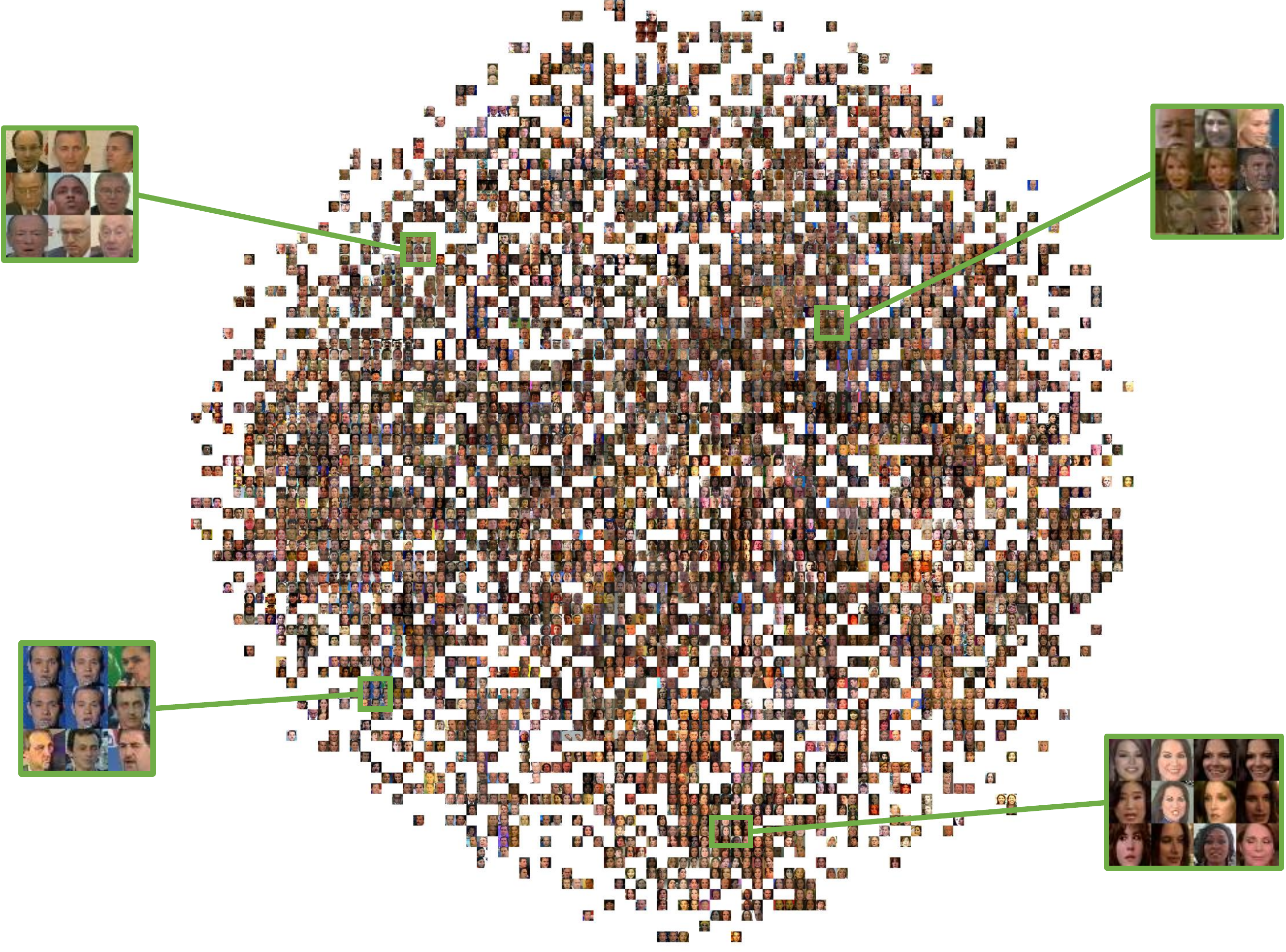}
}
\subfigure[t-SNE visualization of FaceScrub dataset.]{
\includegraphics[width=0.99\linewidth]{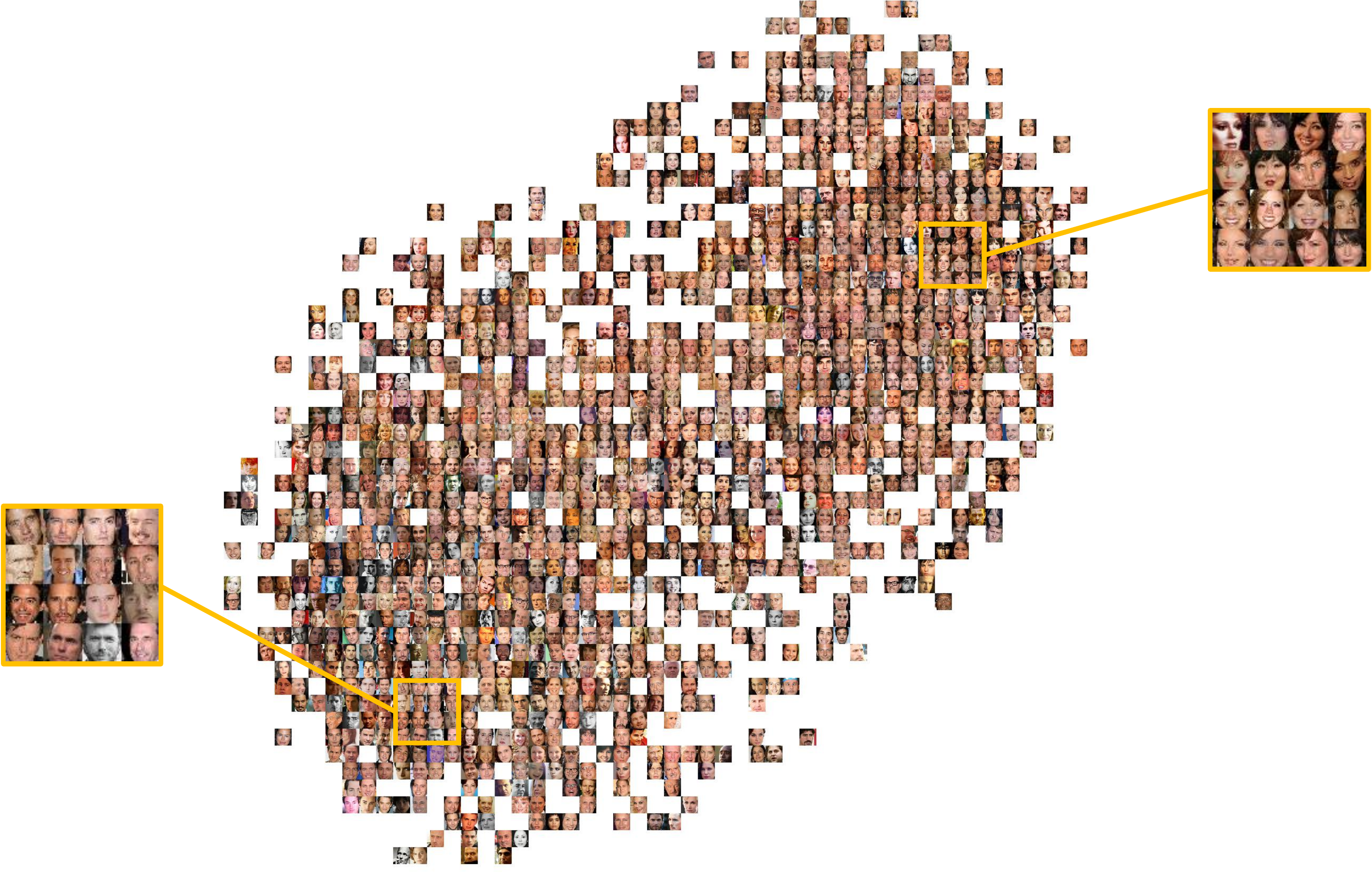}
}
\subfigure[t-SNE visualization of VGGFace2 closed-set dataset.]{
\includegraphics[width=0.99\linewidth]{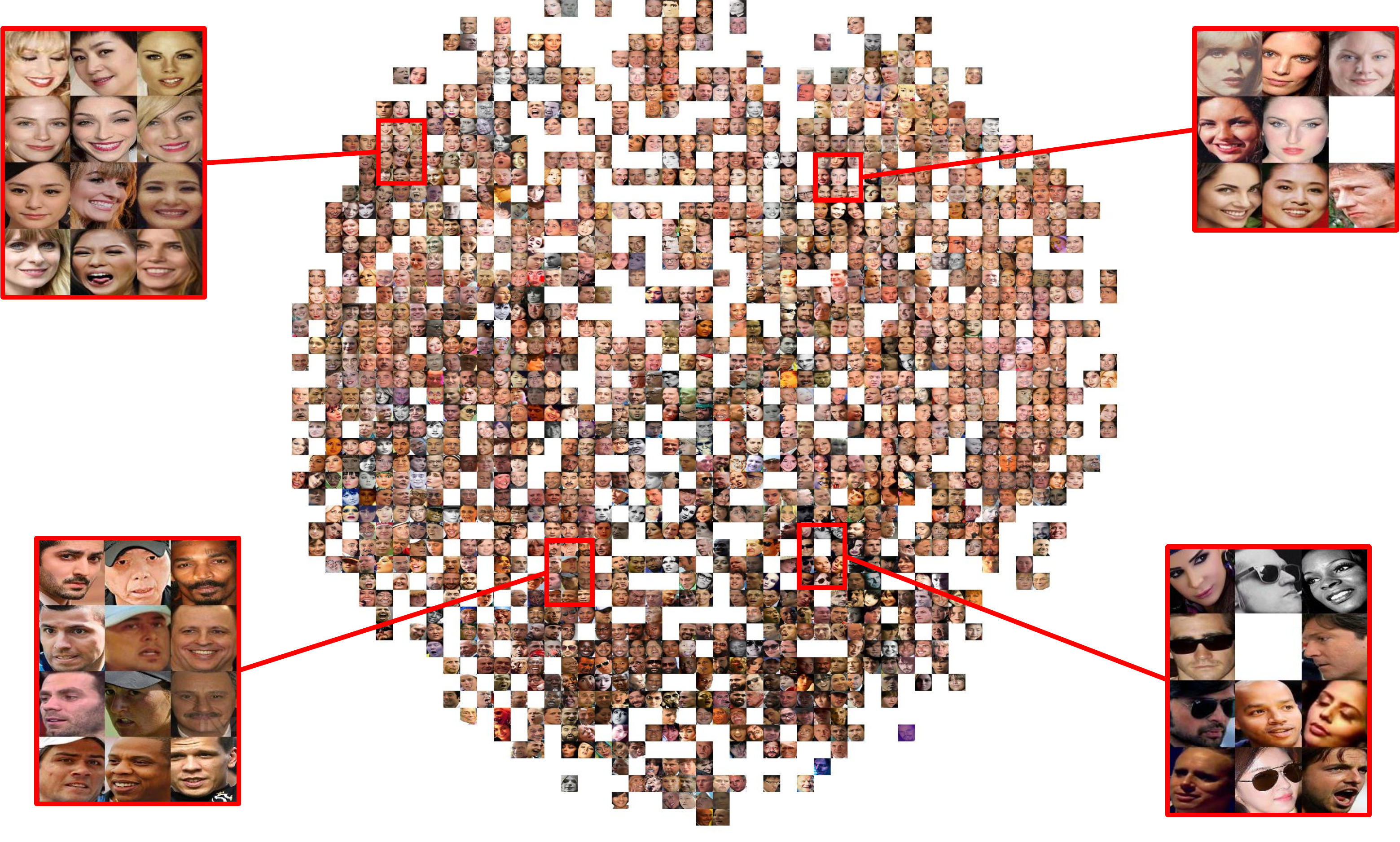}
}
\caption{Retrieval results on face image datasets @48-bits.}

\label{fig:Figure9}
\end{figure}

\subsection{Visualization}

As a qualitative results, we provide actual retrieval results on Figure \ref{fig:Figure8}. From the retrieved images, we are able to find out that even though the query image is gray, the color images of the same identity are listed on the top ranked. In addition, we illustrate t-SNE \cite{t-SNE} visualization results of query (test) images @48-bits in Figure \ref{fig:Figure8}. Even if we do not utilize detailed annotations such as race, gender, beard, or wearing glasses, we can observe that queries containing similar facial attributes are gathered nearby.

\section{Conclusion}
\label{sec:5}

In this paper, we have proposed Similarity Guided Hashing (SGH) network for face image retrieval, which exploits an end-to-end supervised learning strategy. With the randomly transformed face images, we learned the self and pairwise-similarity between the original image and the transformed one in the latent space to find better image representations. In addition, we employed quantization and classification training objectives on hashing head to appropriately encode learned representations into the hash space while minimizing the information loss. In the end, we could generate discriminative binary-like hash codes to perform fast and accurate image retrieval. Retrieval results on the large scale face image datasets with various resolutions verify the effectiveness of our approach with the state-of-the-art performances. We will release our code and the new high-resolution dataset for further research and comparisons.

\newpage

{\small
\bibliographystyle{ieee_fullname}
\bibliography{egbib}
}

\begin{IEEEbiography}{Michael Shell}
Biography text here.
\end{IEEEbiography}

\begin{IEEEbiographynophoto}{John Doe}
Biography text here.
\end{IEEEbiographynophoto}

\end{document}